%% file: main.tex
\documentclass{article}
\usepackage[T1]{fontenc}
\usepackage[utf8]{inputenc}
\usepackage{microtype}
\usepackage{graphicx}
\usepackage{comment}
\usepackage{booktabs,array}
\usepackage{hyperref}

\usepackage[preprint]{icml2026}
\usepackage{amsmath,amssymb,mathtools,amsthm,bm}
\usepackage{tikz}
\usepackage{xcolor}
\newcommand{\grayit}[1]{\textcolor{gray}{\textit{#1}}}
\usepackage{booktabs}
\usepackage{multirow}
\usepackage{graphicx}
\usepackage{svg}
\usepackage{tabularx, booktabs, array}
\usetikzlibrary{arrows.meta,positioning,calc}
\usepackage{pifont}
\newcommand{\cmark}{\ding{51}} 
\newcommand{\xmark}{\ding{55}} 
\usepackage[capitalize,noabbrev]{cleveref}
\definecolor{readblue}{HTML}{236B8E}
\definecolor{writegreen}{HTML}{48734C}
\definecolor{draftcolor}{HTML}{9A4B16}

\newcommand{\method}{GHC}

\theoremstyle{plain}

\theoremstyle{remark}

\icmltitlerunning{\method: Expanding Residual Interfaces in Hyper-Connections}
\hypersetup{pdftitle={\method: Expanding Residual Interfaces in Hyper-Connections},pdfauthor={Anonymous Authors}}

\usepackage{amsmath,amssymb,amsthm}

\theoremstyle{plain}
\newtheorem{theorem}{Theorem}[section]

\theoremstyle{definition}
\newtheorem{assumption}[theorem]{Assumption}

\begin{document}
\twocolumn[
\icmltitle{\method: Expanding Residual Interfaces in Hyper-Connections}
\icmlsetsymbol{equal}{*}
\icmlsetsymbol{corresponding}{$\dagger$}
\begin{icmlauthorlist}
  \icmlauthor{Tianyu Li}{equal,thu}
  \icmlauthor{Hongtao Zhang}{equal,UCAS}
  \icmlauthor{Dongchen Han}{thu}
  \icmlauthor{Jingshu Liu}{thu}
  \icmlauthor{Bojun Wang}{StepFun}
  \icmlauthor{Chen Jiang}{StepFun}
  \icmlauthor{Yulong Huang}{HKUSTGZ}
  \icmlauthor{Wei Chen}{UCAS}
  \icmlauthor{Tianyu He}{corresponding,StepFun}
  \icmlauthor{Xiangyu Zhang}{StepFun}
  \icmlauthor{Gao Huang}{corresponding,thu}
  
\end{icmlauthorlist}

\icmlaffiliation{thu}{
  Tsinghua University
}
\icmlaffiliation{HKUSTGZ}{The Hong Kong University of Science and Technology (Guangzhou)}
\icmlaffiliation{UCAS}{University of Chinese Academy of Sciences}
\icmlaffiliation{StepFun}{StepFun}
\icmlcorrespondingauthor{Gao Huang}{gaohuang@tsinghua.edu.cn}

\icmlcorrespondingauthor{Tianyu He}{hetianyu@stepfun.com}
\icmlkeywords{Residual connections, hyper-connections, Transformers, language modeling, optimization}
\vskip 0.3in
]
\printAffiliationsAndNotice{%
  \icmlEqualContribution\quad
  \textsuperscript{$\dagger$}Corresponding authors.%
}

\begin{figure*}[t]
 \centering
 \includegraphics[width=\linewidth]{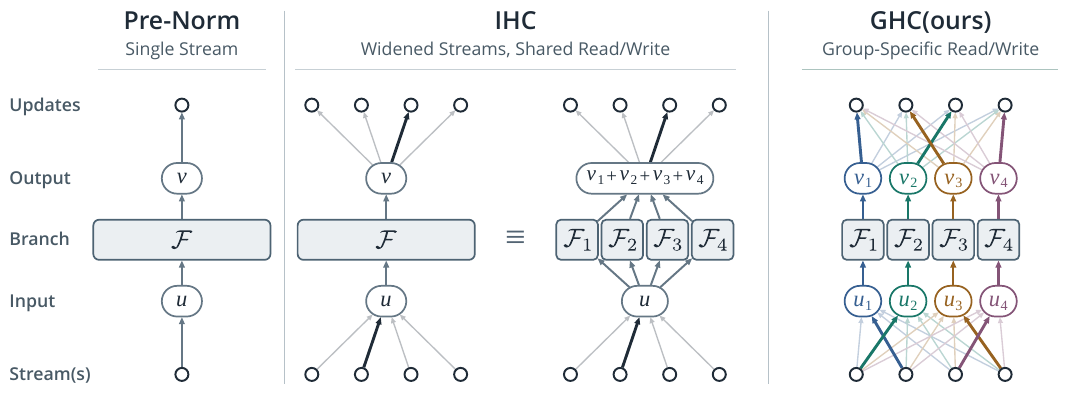}
 \caption{Residual branch reads and writes in Pre-Norm, IHC, and GHC. In IHC, all computation groups read the same \(c\)-dimensional mixture of residual streams, and their outputs are summed before being written back. In GHC, each group reads its own mixture, and each stream combines the group outputs with its own weights. The top row shows each stream’s update from the residual branch.}

 \label{fig:fig1}
\end{figure*}
\begin{abstract}
Hyper-Connections expand residual capacity across parallel streams, yet restrict every branch to narrow read and write interfaces: each branch still reads a single stream mixture and writes scaled copies of one output vector to all streams.
To lift this bottleneck, we propose Grouped Hyper-Connections (GHC).
By partitioning each branch into parallel groups along attention heads or FFN channels, GHC supplies each group with a distinct stream mixture and aggregates their separate outputs into each stream with learned weights.
Extensive experiments across dense and Mixture-of-Experts (MoE) models show that GHC consistently improves upon Identity Hyper-Connections (IHC) with negligible parameter and memory-access overhead, increasing training efficiency by 35\% over Pre-Norm and 10\% over IHC.
Moreover, expanding residual interfaces overcomes the diminishing returns of adding streams, allowing wider residual states to yield further gains.
\end{abstract}

\input{secs/intro}
\input{secs/relatedwork}

\input{secs/preliminaries}
\input{secs/method}

\input{secs/exp}

\input{secs/analysis}

\section{Conclusion}
\label{sec:conclusion}

We introduced \method{}, which expands the read and write interfaces of Hyper-Connections without significantly increasing FLOPs or memory traffic.
By partitioning each branch into parallel groups with width-calibrated $\operatorname{TanhSiLU}$ gates, \method{} enables group-specific signed stream reads and weighted write-backs.
Across scaling ladders and MoE models up to 13B, \method{} consistently improves training efficiency and downstream performance over Pre-Norm and IHC while overcoming diminishing returns of adding streams. 


\bibliography{references}
\bibliographystyle{icml2026}

\clearpage

\input{secs/appendix}
\end{document}

%% file: secs/intro.tex
\section{Introduction}

Residual connections~\citep{resnet}, formulated as $x_{\ell+1} = x_\ell + \mathcal F_\ell(x_\ell)$, form the backbone of modern large language models.
This formulation provides a direct identity path for propagating residual outputs to deeper layers.
However, it accumulates all branch outputs with unit weights into a single fixed-width hidden state.
As depth increases, this uniform accumulation can dilute the relative contribution of later branches and make useful information from earlier layers increasingly difficult to retrieve selectively~\citep{siamesenorm,attentionresiduals}.
From an information-bottleneck perspective, the outputs of $\ell$ preceding residual branches, each of width $c$, are compressed into a single $c$-dimensional hidden state $x_\ell$~\citep{densenet}.
Recent work alleviates this bottleneck by expanding residual capacity and replacing uniform accumulation with learned mixing.
Hyper-Connections (HC)~\citep{hyperconnections} expands the residual state into multiple streams and allows branches to dynamically read different mixtures of them, while Attention Residuals~\citep{attentionresiduals} uses an attention mechanism to selectively aggregate representations across depth.

Despite their different aggregation strategies, HC and Attention Residuals both exhibit diminishing returns from further expansion.
Scaling HC beyond four streams~\citep{hyperconnections,mhc,xhc} or moving from Block to Full Attention Residuals~\citep{attentionresiduals} substantially increases memory traffic for only modest performance gains.
We trace this limitation to a branch's \emph{residual interfaces}, the pathways through which it reads from and writes to the residual state.
As illustrated by Identity Hyper-Connections (IHC) in Figure~\ref{fig:fig1} (middle), a four-stream architecture still routes every branch through a $4 \to 1$ read bottleneck and a $1 \to 4$ write bottleneck: four streams collapse into a single $c$-dimensional input $u$ shared across the entire branch, and a single $c$-dimensional output $v$ is scaled back to all four streams.
Because both interfaces remain $c$-dimensional regardless of the stream count, widening the residual state alone cannot widen a branch's access to it. This raises a natural question:

\emph{Do constrained read and write interfaces limit further performance gains from hyper-connections?}

To thoroughly investigate this question, we introduce \textbf{Grouped Hyper-Connections (\method)}, which partitions each residual branch into $r$ parallel computation groups (Figure~\ref{fig:fig1}, right).
Each group reads a distinct $c$-dimensional linear combination of the residual streams, processes it using a fraction of the original intermediate width, and produces a $c$-dimensional output.
Each residual stream then linearly combines these group outputs using its own learned weights.
By processing multiple input mixtures and generating distinct outputs in parallel, \method{} expands both the read and write interfaces from width $c$ to $rc$.
Crucially, by exploiting the linearity of the branch's input and output projections, \method{} applies seamlessly to Attention, dense FFN, and Mixture-of-Experts (MoE) layers, preserving total intermediate width, nonlinear computation, and branch projection FLOPs with negligible gating overhead.

Extensive experiments answer our question in the affirmative.
Across 430M dense models, a compute-matched scaling ladder, and MoE models up to 13B parameters, \method{} consistently outperforms Pre-Norm and IHC, improving net training efficiency by $35\%$ over Pre-Norm and $10\%$ over IHC while introducing no additional forward memory traffic under proper kernel fusion.
Moreover, whereas IHC stagnates when scaled from four to eight streams, \method{} unlocks substantial further loss reductions from wider residual states, confirming that expanded residual interfaces are key to utilizing multi-stream capacity.

%% file: secs/relatedwork.tex
\section{Related Work}
\label{sec:ghc-related-work}
\noindent\textbf{Depth-wise residual aggregation.}
Beyond uniform residual accumulation, Highway Networks~\citep{highwaynetworks} introduce learned gating across depth, and DenseNet~\citep{densenet} concatenates all preceding feature maps within a block.
In Transformers, DLCL~\citep{dlcl} and DenseFormer~\citep{denseformer} generalize scalar layer mixing~\citep{elmo} to learn static depth-wise weights after each sublayer, while Attention Residuals~\citep{attentionresiduals} computes token-dependent attention over branch outputs or block summaries. Multi-Head Attention Residuals~\citep{mhar} assigns independent
depth-attention distributions to different channel subspaces.
MUDDFormer~\citep{muddformer} constructs separate historical mixtures for the query, key, value, and residual paths, and M3Former~\citep{m3former} further specializes these mixtures across channel groups.
Multi-Stream Attention Residuals~\citep{MSAR} retrieves from multiple temporal views of past layers before merging them into a single sublayer input.
Instead of growing a depth-wise history cache, \method{} maintains a fixed number of residual streams while expanding each branch's read and write interfaces.

\noindent\textbf{Multi-stream residual architectures.}
Maintaining parallel representation streams with learned cross-stream linear mixing echoes early multi-stream vision backbones~\citep{CrossStitchNetworks}.
In language models, Hyper-Connections (HC)~\citep{hyperconnections} widens the residual state into parallel streams, yet each branch still reads a single stream mixture and writes scaled copies of one output vector across streams.
Subsequent variants refine stream propagation and feature slicing: mHC~\citep{mhc} constrains cross-stream transitions to be doubly stochastic, whereas IHC~\citep{hy4preview} and Gated Residual~\citep{qwen3p8next} adopt identity stream shortcuts, with the latter introducing channel-wise read gates. SimpleHC~\citep{simplehc} retains learned read/write gates with
identity stream transitions and reduces repeated state access
through batched reads and deferred writes.
SiameseNorm~\citep{siamesenorm} routes each branch output into dual streams with coupled pre- and post-normalization.
Along the channel dimension, Frac-Connections~\citep{fracconnections} mixes slices of a fixed-width state, StepsNet~\citep{stepsnet} progressively appends new channel groups across stages, and Virtual Width Networks~\citep{vwn} assemble slices of a wider state into a single backbone-width input and write back output slices.
On the write path, xHC~\citep{xhc} augments FFN outputs with convolutions.
Unlike these designs, which keep the residual branch monolithic, \method{} partitions the branch itself to widen both read and write interfaces.

\noindent\textbf{Grouped branch computation.}
Partitioning a transformation into parallel groups has been widely explored for representational cardinality and distributed efficiency.
ResNeXt~\citep{resnext} sums parallel transformations of a shared input, and CondenseNet~\citep{condensenet} learns group-wise input channel pruning across convolutional stages.
In Transformers, multi-head attention~\citep{attention} and tensor parallelism in Megatron-LM~\citep{megatronLM} similarly partition attention heads and FFN intermediate channels, summing partial outputs computed from a shared input, while fine-grained MoE architectures~\citep{deepseekmoe,deepseekv3} route tokens across partitioned FFN experts.
Residual Matrix Transformers~\citep{residual_matrix_transformer} equip each attention head with a dedicated read and outer-product write over a matrix state while concatenating reads in the FFN.
In contrast, \method{} supplies each head, FFN channel group, or routed expert with dedicated stream read and write weights while preserving total intermediate width.

\noindent\textbf{Concurrent work.}
While preparing this manuscript, we became aware of concurrent work studying grouped updates for tensor parallelism in dense Transformers~\citep{tphc}.
Beyond this shared perspective, \method{} introduces signed, calibrated \textnormal{TanhSiLU} gating that unlocks group scaling, extends grouped interfaces to sparse MoE models up to 13B parameters, and demonstrates consistent training efficiency gains across scaling ladders and stream counts.
We further show that \method{} incurs no extra forward memory traffic in principle under operator fusion, and validate its practical inference efficiency in Appendix~\ref{app:kernel_fusion}.

%% file: secs/preliminaries.tex
\section{Preliminaries}
\label{sec:Preliminaries}

\subsection{Residual Architectures}
\label{sec:residual_architectures}

We index attention and FFN branches sequentially by $\ell$
and denote the hidden dimension by $c$.
Let $\mathcal{F}_\ell$ denote the $\ell$-th residual branch, including
its input normalization.
For notational simplicity, we write per-token states as row vectors and
reserve $e \in \mathbb{R}^{1 \times c}$ for the initial token embedding.

\noindent\textbf{Standard residual connections.}
A standard Pre-Norm residual architecture maintains a single hidden stream:
\begin{equation}
    x_{\ell+1} = x_\ell + \mathcal{F}_\ell(x_\ell),
    \label{eq:standard_residual}
\end{equation}
where $x_1 = e$.
Each branch reads the accumulated state and adds its output
back to the same stream.

\noindent\textbf{Hyper-Connections.}
Hyper-Connections (HC)~\citep{hyperconnections} expands
the residual state into $n$ parallel streams,
$X_\ell \in \mathbb{R}^{n \times c}$:
\begin{equation}
    X_{\ell+1}
    = H_\ell^{\mathrm{res}} X_\ell
    + (h_\ell^{\mathrm{post}})^\top
    \mathcal{F}_\ell(h_\ell^{\mathrm{pre}} X_\ell).
    \label{eq:hc}
\end{equation}
Here, the initial state is $X_1 = \mathbf{1}_n e$
with $\mathbf{1}_n \in \mathbb{R}^{n \times 1}$, while
$h_\ell^{\mathrm{pre}}, h_\ell^{\mathrm{post}}
\in \mathbb{R}^{1 \times n}$
aggregate the streams into the branch input and distribute
the branch output back to the streams, respectively.
The transition matrix
$H_\ell^{\mathrm{res}} \in \mathbb{R}^{n \times n}$
governs cross-stream residual mixing.

In dynamic HC, $h_\ell^{\mathrm{pre}}$, $h_\ell^{\mathrm{post}}$, and $H_\ell^{\mathrm{res}}$ are predicted from the flattened,
normalized residual state via linear projections followed by
nonlinear activations. Because $n$ is small (typically $n \le 4$), these predictors require only $\mathcal{O}(n^3 c)$ parameters, which is negligible compared with the $\mathcal{O}(c^2)$ parameters of the residual branch.

Recently, mHC~\citep{mhc} constrains $H_\ell^{\mathrm{res}}$
to be doubly stochastic to stabilize its composition across depth.
We refer to the simplified variant with $H_\ell^{\mathrm{res}} = I_n$
as Identity Hyper-Connections (IHC).
Compared with mHC, IHC achieves comparable performance with
lower system overhead, motivating its adoption in several recent
large language models~\citep{hy4preview, qwen3p8next}.

\subsection{Information Bottlenecks in Residual Interfaces}
\label{sec:interface_bottlenecks}

Although HC widens the residual state to $nc$, it inherits the
single-stream branch design of standard Pre-Norm:
each branch still operates as a monolithic unit with narrow \emph{residual interfaces}, reading a single $c$-dimensional input and writing a single $c$-dimensional output.
To see where this bottleneck arises, we decompose a residual branch
$\mathcal{F}_\ell$ as
\begin{equation}
    \mathcal{F}_\ell(u)
    =
    \Phi_\ell\!\left(
        \operatorname{Norm}(u) W_\ell^{\mathrm{in}}
    \right)
    W_\ell^{\mathrm{out}},
    \label{eq:branch_decomposition}
\end{equation}
where
$W_\ell^{\mathrm{in}} \in \mathbb{R}^{c \times M_1}$ and
$W_\ell^{\mathrm{out}} \in \mathbb{R}^{M_2 \times c}$
are linear projections, and $\Phi_\ell$ encapsulates the intermediate
nonlinear computation (including cross-token interaction in attention).
Specifically, $W_\ell^{\mathrm{in}}$ stacks the query, key, value, and optional gate projections in attention, or the gate and up projections in a gated FFN;
consequently, the intermediate dimensions $M_1$ and $M_2$ need not be equal.

Under HC, the feature dimensions along the branch pathway evolve as
\begin{equation}
    nc
    \xrightarrow{h_\ell^{\mathrm{pre}}} c
    \xrightarrow{W_\ell^{\mathrm{in}}} M_1
    \xrightarrow{\Phi_\ell} M_2
    \xrightarrow{W_\ell^{\mathrm{out}}} c
    \xrightarrow{(h_\ell^{\mathrm{post}})^\top} nc.
    \label{eq:interface_pathway}
\end{equation}
At the input interface, all $n$ streams are collapsed into a single $c$-dimensional vector via a
scalar-weighted sum prior to projection, forcing every component of $\Phi_\ell$ to process the exact same stream mixture. Increasing the intermediate width $M_1$ cannot recover stream-level distinctions already lost during this pre-aggregation. At the output interface, the $M_2$-dimensional intermediate
representation is compressed back into a single $c$-dimensional vector
before being distributed across the $n$ streams.
Consequently, every stream receives only a scaled copy of the same branch output, preventing the branch from delivering stream-specific updates.

Together, these $c$-dimensional interfaces restrict how branch computation
accesses and updates the expanded residual state, underutilizing
the richer representations.
Next, we show how expanding both interfaces unlocks this capacity without widening the intermediate projections.

%% file: secs/method.tex
\section{Expanding Residual Interfaces}
\label{sec:method}

\subsection{Grouped Read, Compute, and Write}
We now introduce \textbf{Grouped Hyper-Connections (\method)}, which partitions each branch's input and output projections into $r$ parallel computation groups along the attention-head or FFN intermediate-channel dimension.
For equally sized groups, the input and output projections of group $i$ have shapes
\begin{equation}
    W_{\ell,i}^{\mathrm{in}}
    \in \mathbb{R}^{c\times M_1/r},
    \qquad
    W_{\ell,i}^{\mathrm{out}}
    \in \mathbb{R}^{M_2/r\times c}.
\end{equation}
Together with its corresponding slice of intermediate computation and group-wise input normalization, these projections define a group function $\mathcal{F}_{\ell,i}$ mapping a $c$-dimensional input to a $c$-dimensional output.
Let
\begin{equation}
    H_\ell^{\mathrm{pre}}(X_\ell),
    H_\ell^{\mathrm{post}}(X_\ell)
    \in\mathbb{R}^{r\times n}
\end{equation}
denote the learned, input-dependent coefficient matrices for stream reading and writing, respectively.
The resulting layer update is given by
\begin{equation}
    \begin{aligned}
        U_\ell
        &= H_\ell^{\mathrm{pre}}(X_\ell)X_\ell, \\
        o_{\ell,i}
        &= \mathcal{F}_{\ell,i}(u_{\ell,i}),
        \qquad i=1,\ldots,r, \\
        O_\ell
        &= [o_{\ell,1};\ldots;o_{\ell,r}], \\
        X_{\ell+1}
        &= H_\ell^{\mathrm{res}}X_\ell
        + \bigl(H_\ell^{\mathrm{post}}(X_\ell)\bigr)^\top O_\ell,
    \end{aligned}
    \label{eq:expanded}
\end{equation}
where $u_{\ell,i} \in \mathbb{R}^{1\times c}$ is the $i$-th row of $U_\ell\in\mathbb{R}^{r\times c}$, and $O_\ell \in \mathbb{R}^{r\times c}$ stacks the group outputs vertically.
Consequently, the feature dimensions along the branch pathway evolve as
\begin{equation}
    nc \xrightarrow{H_\ell^{\mathrm{pre}}} rc \xrightarrow{W_\ell^{\mathrm{in}}} M_1
    \xrightarrow{\Phi_\ell} M_2 \xrightarrow{W_\ell^{\mathrm{out}}} rc \xrightarrow{(H_\ell^{\mathrm{post}})^\top} nc.
    \label{eq:expanded_pathway}
\end{equation}
This construction expands both residual interfaces from $c$ to $rc$ while preserving the total intermediate widths $M_1$ and $M_2$.
Because the groups partition the existing projections, their combined projection FLOPs remain unchanged.
Although the interface operations remain linear combinations over streams, they now supply distinct inputs to each computation group and keep group outputs separate until write-back mixing.
\Cref{fig:interface} in Appendix~\ref{app:architecture_diagram} illustrates this construction for $n=r=4$ and $H_\ell^{\mathrm{res}}=I_n$.

\subsection{Properties of Interface Expansion}

\noindent\textbf{Preserving nonlinearity.}
Our grouping strategy exploits the inherently separable structure of the intermediate nonlinear computation $\Phi_\ell$ in Eq.~\eqref{eq:branch_decomposition}.
In multi-head attention, attention logits, softmax normalization, and value aggregation operate independently within each head; in dense FFNs, nonlinear activations and multiplicative gating act elementwise across intermediate channels before being combined linearly by the output projection.
Partitioning along heads or intermediate channels therefore leaves all nonlinear operations and their total intermediate width completely intact.
Each group retains full access to all $c$ input channels and projects back to all $c$ output channels, while the expanded interfaces equip these parallel sub-computations with dedicated read and write pathways.

\noindent\textbf{Recovering IHC.}
When all rows of $H_\ell^{\mathrm{pre}}$ and $H_\ell^{\mathrm{post}}$ are tied to the single-row IHC vectors $h_\ell^{\mathrm{pre}}$ and $h_\ell^{\mathrm{post}}$, every group receives an identical input and distributes its output with identical weights.
By the linearity of the output projection, the partitioned branches then satisfy $\sum_{i=1}^{r}\mathcal{F}_{\ell,i}(u)=\mathcal{F}_\ell(u)$.
Setting $H_\ell^{\mathrm{res}}=I_n$ thus  recovers IHC as a special case.

\begin{table*}[t]
    \centering
    \caption{
        Results on dense and Mixture-of-Experts (MoE) models.
        Lower is better for average loss and LAMBADA perplexity; all other metrics are reported as percentages.
        Avg.\ Score is the unweighted mean of the six displayed two-decimal task scores from PIQA to HumanEval.
        Bold indicates the best result within each model-size group, including ties.
    }
    \label{tab:main_results}
    \small
    \setlength{\tabcolsep}{4pt}
    \renewcommand{\arraystretch}{1.12}
    \resizebox{\textwidth}{!}{%
    \begin{tabular}{@{}ll*{9}{c}@{}}
        \toprule
        Model & Method
        & Avg.\ Loss & LAMBADA$\downarrow$
        & PIQA & HellaSwag & TriviaQA
        & MATH & GSM8K & HumanEval & Avg.\ Score \\
        \midrule

        \multirow{3}{*}{430M Dense}
        & Pre-Norm
        & 2.0713 & 14.41 & 67.19 & 47.44 & 13.64 & 3.54 & 2.96 & 7.32 & 23.68 \\
        & IHC
        & 2.0351 & \textbf{12.29} & 68.01 & 49.75 & 16.09 & 3.56 & 2.73 & \textbf{10.37} & 25.09 \\
        & \textbf{GHC(ours)}
        & \textbf{2.0161} & 12.47 & \textbf{68.77} & \textbf{50.79} & \textbf{16.94} & \textbf{4.06} & \textbf{3.87} & \textbf{10.37} & \textbf{25.80} \\

        \midrule
        \multirow{3}{*}{4BA0.2B MoE}
        & Pre-Norm
        & 1.9260 & 12.95 & 71.55 & 54.68 & 23.31 & 5.52 & 5.84 & 12.20 & 28.85 \\
        & IHC
        & 1.9066 & 10.67 & 71.87 & 55.98 & 24.90 & 6.14 & 5.31 & 14.02 & 29.70 \\
        & \textbf{GHC(ours)}
        & \textbf{1.8862} & \textbf{10.30} & \textbf{71.98} & \textbf{57.83} & \textbf{26.78} & \textbf{7.40} & \textbf{5.99} & \textbf{17.68} & \textbf{31.28} \\

        \midrule
        \multirow{3}{*}{13BA0.6B MoE}
        & Pre-Norm
        & 1.7052 & 5.72 & 76.17 & 67.96 & 43.30 & 15.10 & 21.68 & 23.78 & 41.33 \\
        & IHC
        & 1.6901 & 5.69 & \textbf{78.18} & 69.10 & 45.48 & 15.80 & 21.99 & 25.00 & 42.59 \\
        & \textbf{GHC(ours)}
        & \textbf{1.6775} & \textbf{5.16} & 77.64 & \textbf{69.91} & \textbf{48.59} & \textbf{17.36} & \textbf{24.03} & \textbf{26.22} & \textbf{43.96} \\

        \bottomrule
    \end{tabular}
    }
\end{table*}

\subsection{Adaptation to Mixture-of-Experts}
\label{subsec:method}

Unlike attention heads and dense FFN channels, the top-$k$ routing mechanism in a Mixture-of-Experts (MoE) layer is not separable across groups, even though its intermediate computation is naturally partitioned across experts.
Forcibly splitting the router into independent sub-routers (e.g., replacing global top-$8$ routing over $256$ experts with eight separate top-$1$ routers over $32$ experts each) alters the standard MoE formulation and degrades performance.

To preserve standard global top-$k$ competition while still widening residual access, we decouple routing from expert-level stream mixing.
Observing that routing compute scales linearly rather than quadratically with the hidden dimension $c$, we retain a unified router conditioned on the concatenated residual streams, expanding the router projection from $c \to E$ to $nc \to E$ (e.g., $4c \to 256$ for $n=4$ and $E=256$ experts).
Furthermore, because distributed MoE training is sensitive to cross-device communication overhead, we omit token-dependent dynamic gating at the expert level and instead assign each expert $2n$ learnable static parameters (passed through the gate activation introduced next) as its dedicated read and write coefficients across the $n$ streams.
We ablate these MoE design choices in detail in Appendix~\ref{app:moe_ablation}.
\subsection{Calibrating Dynamic Read and Write Gates}
\label{sec:gate_calibration}
Under the dynamic HC paradigm~\citep{hyperconnections,mhc}, the read and write maps $H_\ell^{\mathrm{pre}}$ and $H_\ell^{\mathrm{post}}$ use token-dependent mixing coefficients shared across $c$ hidden channels.
Specifically, each predictor $\star \in \{\mathrm{pre}, \mathrm{post}\}$ takes the flattened, normalized residual state $z_\ell \in \mathbb{R}^{nc}$ and computes $\phi(\alpha_\ell^\star W_\ell^\star z_\ell + b_\ell^\star)$, with learnable weights $W_\ell^\star$, scalar scale $\alpha_\ell^\star$, bias $b_\ell^\star$, and elementwise activation $\phi$. Because this parameterization governs both residual expressiveness and gradient dynamics (see Appendix~\ref{app:gate_calibration} for a comparison of existing designs), we calibrate two key components: the activation $\phi$ and the predictor's width scaling.

\noindent\textbf{Signed and asymmetric activations.}
Starting from the sigmoid-gated IHC baseline, we find that replacing sigmoid with either $\tanh$ or the composite activation
\begin{equation}
    \phi(t)
    = \operatorname{TanhSiLU}(t)
    = \tanh\!\left(\operatorname{SiLU}(t)\right),
    \label{eq:gate_activation}
\end{equation} where $\operatorname{SiLU(t)}=t\sigma(t)$,
yields consistent loss reductions and suppresses residual outliers once the first-layer read map is fixed as described below.
In \method{}, these gains grow steadily with the number of computation groups $r$, motivating signed gate activations for expanded residual interfaces.

$\operatorname{TanhSiLU}$ offers two advantages over sigmoid.
First, it provides non-vanishing, well-conditioned local gradients near zero.
For a sigmoid gate $a=\sigma(t)$, the derivative $a(1-a)$ vanishes as $a \to 0$; suppressing a stream therefore drives the gate into saturation, slowing subsequent adaptation.
In contrast, $\phi(0)=0$ and $\phi'(0)=\frac{1}{2}$.
Across small nonnegative gate values in $[0, 0.5]$, $\phi$ maintains relatively uniform sensitivity and can reach or cross zero at a finite preactivation, allowing a group to mute a stream while retaining responsive gradients. Second, signed coefficients unlock more expressive stream combinations.
For a read mixture $u=\sum_b h_b x_b$, nonnegative weights confine $u$ to the convex cone spanned by the streams $\{x_b\}$.
Allowing negative weights lets each group take differences across streams and cancel redundant components, similar in spirit to differential attention~\citep{differentialtrnasformer} and delta-rule updates in linear attention~\citep{schlag2021linear,inline,gateddeltanet}.
The same flexibility applies when writing group outputs back to the streams.
Compared with symmetric $\tanh$, $\operatorname{TanhSiLU}$ has an asymmetric range (approximately $[-0.27, 1)$): it supports full-strength positive aggregation alongside bounded negative corrections, introducing a helpful inductive bias toward additive accumulation with selective subtraction.

However, pairing signed gates with a learnable first-layer read map can trigger gradient-norm spikes and occasional loss instabilities, echoing observations in prior work~\citep{mhc,qwen3p8next}.
Because all initial residual streams hold identical copies of the token embedding $e$, group $i$ in the first attention layer receives $q_i e$ with $q_i=\sum_b (H_1^{\mathrm{pre}})_{ib}$.
Neglecting the normalization $\epsilon$, we have
\begin{equation}
    \operatorname{RMSNorm}(q_i e)
    = \operatorname{sign}(q_i)\operatorname{RMSNorm}(e).
    \label{eq:first_layer_gate_sign}
\end{equation}
Consequently, the normalized input has zero gradient with respect to $H_1^{\mathrm{pre}}$ almost everywhere, yet undergoes a discontinuous sign flip whenever $q_i$ crosses zero.
To eliminate this discontinuity, we \textbf{fix $H_1^{\mathrm{pre}}$} throughout training with positive row sums. Notably, a fixed first-layer read map is also used by~\citet{deepseekv4p1}, albeit not from this perspective. With this modification, loss spikes disappear across all runs.

\noindent\textbf{Width calibration.}
Because each mixing coefficient acts across all $c$ channels while $z_\ell$ has $nc$ coordinates, an unscaled linear predictor causes preactivation updates to grow with width $c$ (see Appendix~\ref{app:gate-width} for a formal analysis).
Moreover, while initializing $\alpha_\ell^\star$ to a small constant (e.g., $0.01$ in HC and mHC~\citep{hyperconnections,mhc}) attenuates gradients to $W_\ell^\star$, it leaves the gradient with respect to $\alpha_\ell^\star$ itself unattenuated~\citep{ohc}.
To ensure width-invariant $\Theta(1)$ preactivation updates without relying on small initial scales, we replace the shared scalar $\alpha_\ell^\star$ with entry-wise learnable scales $s_\ell^\star \in \mathbb{R}^{r \times n}$ and incorporate an explicit $1/\sqrt{c}$ factor:
\begin{equation}
    \begin{gathered}
        H_\ell^\star(X_\ell)
        =
        \operatorname{TanhSiLU}\!\left(
            \frac{s_\ell^\star \odot \operatorname{reshape}(W_\ell^\star z_\ell)}{\sqrt{c}}
            + b_\ell^\star
        \right), \\[4pt]
        \star \in \{\mathrm{pre},\mathrm{post}\},
        \qquad s_{\ell,\mathrm{init}}^\star = 1_{r \times n},
    \end{gathered}
    \label{eq:calibrated_gate}
\end{equation}
where $W_\ell^\star z_\ell$ is reshaped into $\mathbb{R}^{r \times n}$ and $b_\ell^\star \in \mathbb{R}^{r \times n}$ denotes entry-wise biases.
We initialize $b_\ell^{\mathrm{post}} = 0.7969$ ($\operatorname{TanhSiLU}(b_\ell^{\mathrm{post}}) \approx 0.5$), and set $b_\ell^{\mathrm{pre}}$ to $2.537$ on stream $((\ell - 1)r + i - 1) \bmod n + 1$ for group $i$ and $0.035$ elsewhere, giving initial gate values of $0.982$ and $0.018$.
We apply Eq.~\eqref{eq:calibrated_gate} to all dynamic read and write maps, while keeping $H_1^{\mathrm{pre}}$ fixed at this initial cyclic pattern.
\subsection{Infrastructure Efficiency}
\label{sec:infrastructure_efficiency}

Unlike widening external residual streams, \method{} expands only the internal branch interfaces ($rc$) while keeping the external residual state ($nc$) and total intermediate widths ($M_1, M_2$) fixed.
Although a naive PyTorch implementation materializes the $rc$-dimensional group inputs and outputs in global memory, these interfaces sit directly adjacent to the group projections and are amenable to kernel fusion:
\begin{equation}
    \begin{aligned}
        X_\ell
        &\xrightarrow{\text{Read+Norm+Proj}}
        Z_\ell \in \mathbb{R}^{1 \times M_1},\\
        Z_\ell
        &\xrightarrow{\Phi_\ell}
        A_\ell \in \mathbb{R}^{1 \times M_2},\\
        (A_\ell,X_\ell)
        &\xrightarrow{\text{Proj+Write+Add}}
        X_{\ell+1} \in \mathbb{R}^{n \times c},
    \end{aligned}
    \label{eq:fused_memory_path}
\end{equation}
\begin{figure}[t]
    \centering
    \includegraphics[width=\columnwidth]{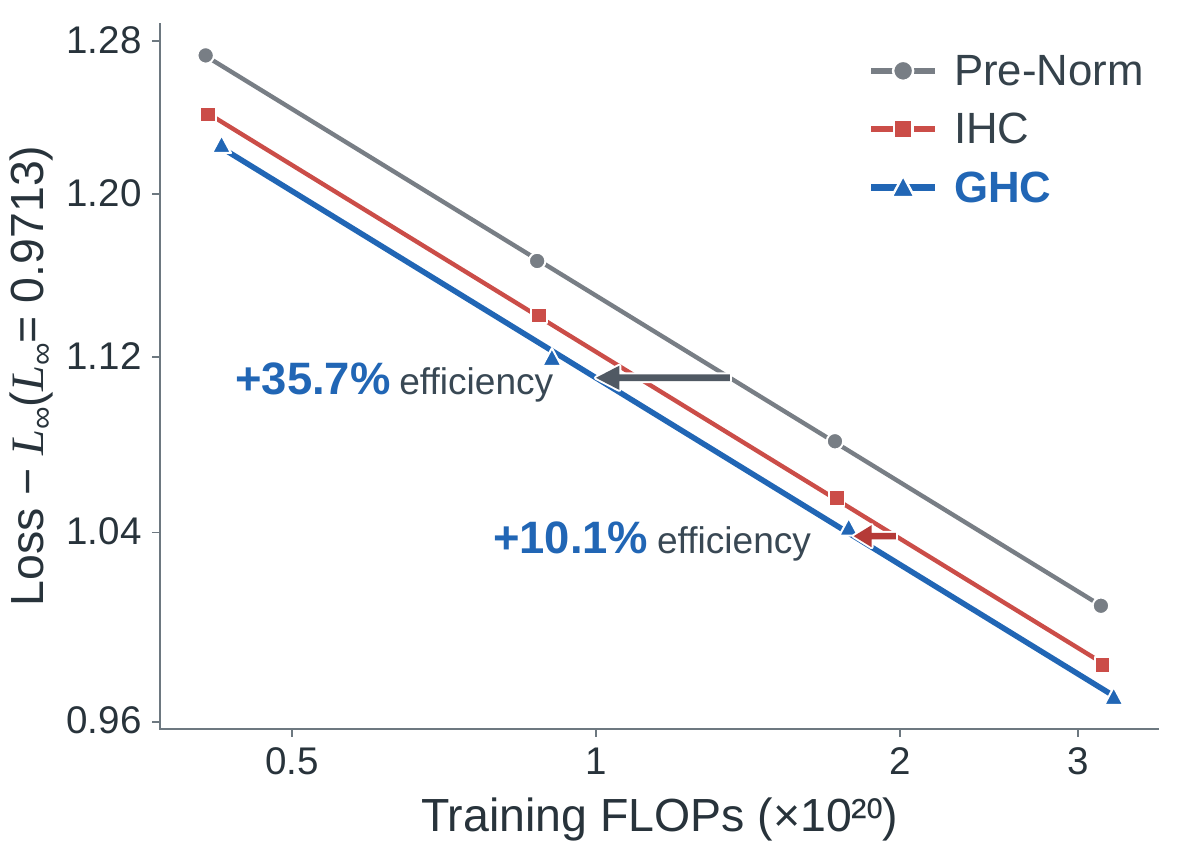}
    \caption{
        Validation loss across the scaling ladder as a function of total training FLOPs for Pre-Norm, IHC, and \method{}.
        Fitted power-law curves indicate that \method{} achieves net training efficiency gains (EG) of approximately $35\%$ over Pre-Norm and $10\%$ over IHC.
    }
    \label{fig:scaling_ladder}
\end{figure}
where each arrow indicates a target fused kernel boundary writing to global memory (given precomputed gate coefficients and normalization statistics), with $Z_\ell$ and $A_\ell$ denoting the projected inputs and outputs of $\Phi_\ell$.
In principle, fusing stream reading with the input projections and stream writing with the output projections allows the $rc$-dimensional intermediate states to stay in on-chip SRAM, matching the forward global-memory traffic of IHC; we present practical operator fusion benchmarks in Appendix~\ref{app:kernel_fusion}.

On the compute side, gate prediction, stream mixing, and group normalization add $\mathcal{O}(rn^2 c)$ FLOPs per token, which is asymptotically small relative to the $\mathcal{O}(c^2)$ branch projections as width $c$ grows; our scaling-ladder experiments (Section~\ref{sec:main_results}) explicitly include these extra FLOPs and verify net compute-efficiency gains.
At the same time, choosing an overly large $r$ increases gating and normalization overhead and shrinks per-group matrix dimensions ($M_1/r, M_2/r$), which can reduce GEMM utilization. Building on the $n=4$ IHC backbone ($H_\ell^{\mathrm{res}}=I_n$), we adopt a moderate group count of $r=8$ as the default \method{} configuration, balancing expressivity and execution overhead. Partitioning branches into eight groups aligns naturally with tensor and expert parallelism without adding pipeline-parallel overhead.

%% file: secs/exp.tex
\section{Experiments}
\label{sec:experiments}

\subsection{Experimental Setup}

Our experiments span three settings (see Appendix~\ref{app:training_details} for full training details).
First, for controlled comparison against competitive baselines, we train a 430M-parameter dense model on 50B tokens.
Second, we evaluate \method{} across a scaling ladder~\citep{scalingladder} to quantify training efficiency gains~\citep{mai1} over Pre-Norm and IHC.
Finally, we scale our evaluation to 4BA0.2B and 13BA0.6B Mixture-of-Experts(MoE) models.

\noindent\textbf{Downstream evaluation.}
Across all models, we evaluate commonsense reasoning, factual knowledge, mathematical reasoning, and code generation using HellaSwag~\citep{hellaswag}, PIQA~\citep{piqa}, TriviaQA~\citep{triviaqa}, MATH~\citep{math}, GSM8K~\citep{gsm8k}, and HumanEval~\citep{humaneval}, and additionally report final-word perplexity on LAMBADA~\citep{lambada}.
We report the unweighted mean of the six percentage-based benchmark scores as Avg.\ Score, excluding average validation loss and LAMBADA perplexity.

\subsection{Main Results}
\label{sec:main_results}\begin{table}[t]
    \centering
    \caption{
        Effect of expanding residual state capacity on the 430M dense model (50B tokens).
        Entries in gray italics denote high-I/O configurations with expanded state access ($n=8$ streams or block size $b=1$ for full depth attention).
        $\Delta$ denotes the loss reduction from widening the residual state within each architecture.
    }
    \label{tab:dense_results}
    \small
    \setlength{\tabcolsep}{0pt}
    \begin{tabular*}{0.85\columnwidth}{@{\extracolsep{\fill}}llcc@{}}
        \toprule
        Method & Setting & Avg.\ Loss $\downarrow$ & $\Delta \uparrow$ \\
        \midrule
        \multirow{2}{*}{IHC}       & $n=4$          & 2.0351                   & \multirow{2}{*}{0.0028} \\
                                   & \grayit{$n=8$} & \grayit{2.0323}          & \\
        \midrule
        \multirow{2}{*}{AttnRes}   & $b=4$          & 2.0363                   & \multirow{2}{*}{0.0041} \\
                                   & \grayit{$b=1$} & \grayit{2.0322}          & \\
        \midrule
        \multirow{2}{*}{\method{}} & $n=4$          & \textbf{2.0161}          & \multirow{2}{*}{\textbf{0.0195}} \\
                                   & \grayit{$n=8$} & \grayit{\textbf{1.9966}} & \\
        \bottomrule
    \end{tabular*}
\end{table}
\noindent\textbf{Breaking the residual interface bottleneck.}
We first evaluate overall model quality on the 430M dense model trained for 50B tokens (Tables~\ref{tab:main_results} and~\ref{tab:dense_results}).
To ensure a fair comparison across multi-stream methods, we apply no specialized boundary projections: all streams are initialized by replicating the token embedding and summed directly before the final normalization and language modeling head.
As shown in Table~\ref{tab:main_results}, under the standard four-stream budget ($n=4$), \method{} ($r=8$) lowers average validation loss to $2.0161$, outperforming Pre-Norm ($2.0713$) by $0.0552$ and IHC ($2.0351$) by $0.0190$, while improving the downstream Avg.\ Score to $25.80$ ($+2.12$ and $+0.71$ points over Pre-Norm and IHC, respectively).
Table~\ref{tab:dense_results} further shows that \method{} ($n=4$) surpasses Block Attention Residuals (AttnRes, block size $b=4$, $2.0363$)~\citep{attentionresiduals}, and even outperforms the high-I/O Full AttnRes variant ($b=1$) by $0.0161$.

Crucially, the paired comparisons in Table~\ref{tab:dense_results} validate our core hypothesis on residual interfaces.
When residual interfaces remain restricted to width $c$, expanding the external residual state yields sharp diminishing returns: doubling IHC from $n=4$ to $n=8$ streams reduces loss by only $0.0028$ ($2.0351 \to 2.0323$), and moving from Block ($b=4$) to Full ($b=1$) AttnRes yields a similarly modest $0.0041$ gain ($2.0363 \to 2.0322$).
Conversely, once \method{} expands the read/write interfaces, scaling from $n=4$ to $n=8$ streams delivers a substantial $0.0195$ loss reduction ($2.0161 \to 1.9966$), nearly $7\times$ the gain of IHC and widening the lead over $8$-stream IHC to $0.0357$.
This confirms that narrow interfaces are the primary bottleneck preventing existing architectures from utilizing wider residual states.

\noindent\textbf{Consistent gains along the scaling ladder.}
To assess how these improvements scale with compute, Figure~\ref{fig:scaling_ladder} plots validation loss against total training FLOPs along the scaling ladder~\citep{scalingladder} for Pre-Norm, IHC, and \method{}.
Across all model sizes and compute budgets, \method{} consistently shifts the scaling curve downward without diminishing returns at larger scales.
Following~\citet{mai1}, we fit power-law curves to quantify the training efficiency gain (EG), defined as the relative compute required by a baseline to match the validation loss of \method{}.
Notably, even with the slight FLOPs overhead from interface mixing and group normalization accounted for in total training FLOPs, \method{} achieves net efficiency gains of approximately $35\%$ over Pre-Norm and $10\%$ over IHC.

\noindent\textbf{Broad improvements on MoE models.}
As shown in Table~\ref{tab:main_results}, the advantages of \method{} extend consistently from dense backbones to sparse MoE architectures.
On the 4B MoE model, \method{} achieves the best performance across every metric, lowering validation loss by $0.0398$ over Pre-Norm and $0.0204$ over IHC while raising the average downstream score to $31.28$ ($+2.43$ and $+1.58$ over Pre-Norm and IHC, respectively).
At the 13B MoE scale, \method{} similarly maintains the lowest validation loss and the highest overall downstream score, with particularly strong gains on mathematical reasoning and coding benchmarks (GSM8K, MATH, and HumanEval).
Notably, expanding the router input across streams adds less than $4\%$ active parameters and $0.2\%$ total parameters, confirming that these benefits stem from wider residual interfaces rather than extra model capacity.

\subsection{Ablation Studies}
\label{sec:ablation}

We ablate four core design choices of \method{} on the 430M dense model with $n=4$:
(1) the gate activation function, (2) the number of computation groups $r$, (3) the placement of expanded interfaces across branches and read/write pathways, and (4) gate width calibration and scale parameterization.
When comparing gate activations, we adjust the initial biases accordingly to ensure identical initial gate values across variants.
Table~\ref{tab:ablation} summarizes all ablation results.

\noindent\textbf{Signed gating unlocks group scaling.}
Comparing sigmoid and $\operatorname{TanhSiLU}$ gating across $r \in \{1, 2, 4, 8\}$ (Table~\ref{tab:ablation}a), $\operatorname{TanhSiLU}$ outperforms sigmoid at every group count, with its validation loss advantage growing monotonically from $0.0034$ at $r=1$ to $0.0106$ at $r=8$.
Over the full range from $r=1$ to $r=8$, $\operatorname{TanhSiLU}$ captures nearly twice the group-scaling reduction of sigmoid.
This widening margin confirms that signed, subtractive mixing and non-vanishing zero-crossing gradients become increasingly beneficial as more groups interact with the residual streams.

\begin{table}[t]
    \centering
    \caption{
        Ablations of \method{} on the 430M dense model ($n=4$).
        \textbf{(a)} Gate activation and group count $r$, where $\Delta$ denotes the loss reduction of $\operatorname{TanhSiLU}$ over sigmoid.
        \textbf{(b)} Interface expansion placement under $\operatorname{TanhSiLU}$, showing $r \in \{1, 8\}$ assigned to the read ($H^{\mathrm{pre}}$) and write ($H^{\mathrm{post}}$) maps of attention and FFN branches.
        \textbf{(c)} Gate scale parameterization and width calibration ($r=8, \operatorname{TanhSiLU}$), reporting peak post-warmup global gradient norm and final validation loss alongside Pre-Norm for reference.
    }
    \label{tab:ablation}
    \small
    \textbf{(a) Gate activation \& group count $r$}\\[2pt]
    \setlength{\tabcolsep}{10pt}
    \begin{tabular}{@{}cccc@{}}
        \toprule
        Groups $r$ & Sigmoid & $\operatorname{TanhSiLU}$ & $\Delta \uparrow$ \\
        \midrule
        1 & 2.0357 & 2.0323 & 0.0034 \\
        2 & 2.0326 & 2.0260 & 0.0066 \\
        4 & 2.0327 & 2.0222 & 0.0105 \\
        8 & 2.0267 & \textbf{2.0161} & \textbf{0.0106} \\
        \bottomrule
    \end{tabular}

    \vspace{6pt}

    \textbf{(b) Interface placement ($\operatorname{TanhSiLU}$)}\\[2pt]
    \setlength{\tabcolsep}{8pt}
    \begin{tabular}{@{}ccccc@{}}
        \toprule
        \multicolumn{2}{c}{Attention} & \multicolumn{2}{c}{FFN} & \multirow{2}{*}{Avg.\ Loss $\downarrow$} \\
        \cmidrule(lr){1-2} \cmidrule(lr){3-4}
        $H^{\mathrm{pre}}$ & $H^{\mathrm{post}}$ & $H^{\mathrm{pre}}$ & $H^{\mathrm{post}}$ & \\
        \midrule
        1 & 1 & 1 & 1 & 2.0323 \\
        8 & 8 & 1 & 1 & 2.0239 \\
        1 & 8 & 1 & 8 & 2.0256 \\
        8 & 8 & 8 & 8 & \textbf{2.0161} \\
        \bottomrule
    \end{tabular}

    \vspace{6pt}

    \textbf{(c) Gate calibration ($r=8, \operatorname{TanhSiLU}$)}\\[2pt]
    \setlength{\tabcolsep}{4.2pt}
    \begin{tabular}{@{}lccc@{}}
        \toprule
        Scale Parameterization & Init.\ Scale & Max Grad $\downarrow$ & Avg.\ Loss $\downarrow$ \\
        \midrule
        \grayit{Pre-Norm baseline}           & \grayit{N/A} & \grayit{0.1262} & \grayit{2.0713} \\
        Scalar $\alpha$, w/o $1/\sqrt{c}$    & $0.01$       & 4.0963          & 2.0192 \\
        Matrix $s$, w/o $1/\sqrt{c}$    & $0.01$       & 0.6780          & 2.0182 \\
        Matrix $s$, w/ $1/\sqrt{c}$ & $1.0$ & \textbf{0.1186} & \textbf{2.0161} \\
        \bottomrule
    \end{tabular}
\end{table}

\begin{figure*}[t]
    \centering
    \includegraphics[width=\linewidth]{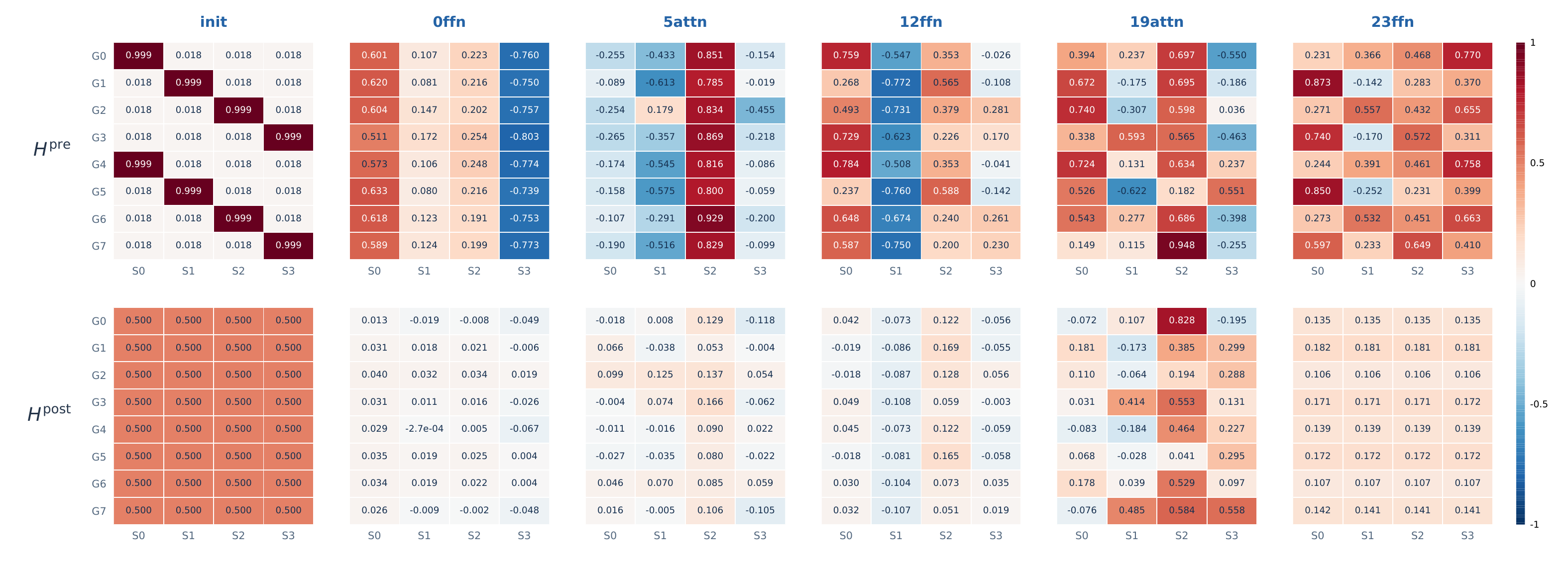}
    \caption{
        Token-averaged read gates $H^{\mathrm{pre}}$ (top, row-wise $\ell_2$-normalized before averaging) and write gates $H^{\mathrm{post}}$ (bottom) in the 430M \method{} model ($n=4, r=8, \operatorname{TanhSiLU}$) at initialization and five trained sublayers.
        Each $8 \times 4$ panel plots the $8$ groups (rows) against the $4$ streams (columns), with red, white, and blue denoting positive, near-zero, and negative values.
    }
    \label{fig:group_specialization}
\end{figure*}
\noindent\textbf{Consistent gains from more computation groups.}
With intermediate widths and projection FLOPs held fixed, increasing $r$ isolates the effect of expanding the branch interface from $c$ to $rc$ (Table~\ref{tab:ablation}a).
Under $\operatorname{TanhSiLU}$, validation loss drops monotonically across $r \in \{1, 2, 4, 8\}$ ($2.0323 \to 2.0260 \to 2.0222 \to 2.0161$), for a total reduction of $0.0162$.
Crucially, scaling from $r=4$ to $r=8$ yields a further $0.0061$ gain even though $r$ exceeds the stream count ($n=4$): because each group shares scalar mixing weights across all $c$ channels, finer partitioning relaxes this coarse group-sharing constraint and lets smaller head or channel subsets specialize on distinct stream mixtures.

\noindent\textbf{Complementary gains across branches and interfaces.}
Table~\ref{tab:ablation}b disentangles the contributions of branch types and read/write pathways between the unexpanded $\operatorname{TanhSiLU}$ baseline ($r=1$, $2.0323$) and \method{} ($r=8$, $2.0161$).
Restricting grouped interfaces to attention ($r=1$ in FFN) yields $2.0239$, capturing $0.0084$ of the gain over the baseline while leaving $0.0078$ from FFN expansion, showing near-equal branch contributions.
Along the interface axis, retaining grouped writes while collapsing $H^{\mathrm{pre}}$ to $r=1$ improves over the baseline by $0.0067$, whereas grouped reads contribute the remaining $0.0095$.
Expanding both branches and interfaces is thus necessary for the full gain.

\noindent\textbf{Width calibration stabilizes optimization.}
Table~\ref{tab:ablation}c evaluates our explicit $1/\sqrt{c}$ gate calibration ($r=8$, $\operatorname{TanhSiLU}$) against uncalibrated variants initialized at scale $0.01$.
Without $1/\sqrt{c}$ scaling, sharing a scalar $\alpha_\ell^\star \in \mathbb{R}$ across all $rn$ gate entries causes severe post-warmup gradient spikes, reaching a peak global gradient norm of $4.1$ ($32\times$ higher than Pre-Norm's $0.126$) as unscaled inner products accumulate gradients onto a single parameter.
Replacing the shared scalar $\alpha_\ell^\star$ with entry-wise scales $s_\ell^\star \in \mathbb{R}^{r \times n}$ lowers the peak post-warmup global gradient norm from $4.10$ to $0.68$ and the validation loss from $2.0192$ to $2.0182$.
The width calibration in Eq.~\eqref{eq:calibrated_gate} introduces an explicit $1/\sqrt{c}$ factor in the gradient with respect to each scale.
With this calibration and all scales initialized to one, \method{} reaches a peak gradient norm of $0.12$, close to the Pre-Norm value of $0.1262$, and achieves the lowest validation loss in this ablation ($2.0161$).

\begin{figure}[t]
    \centering
    \includegraphics[width=\columnwidth]{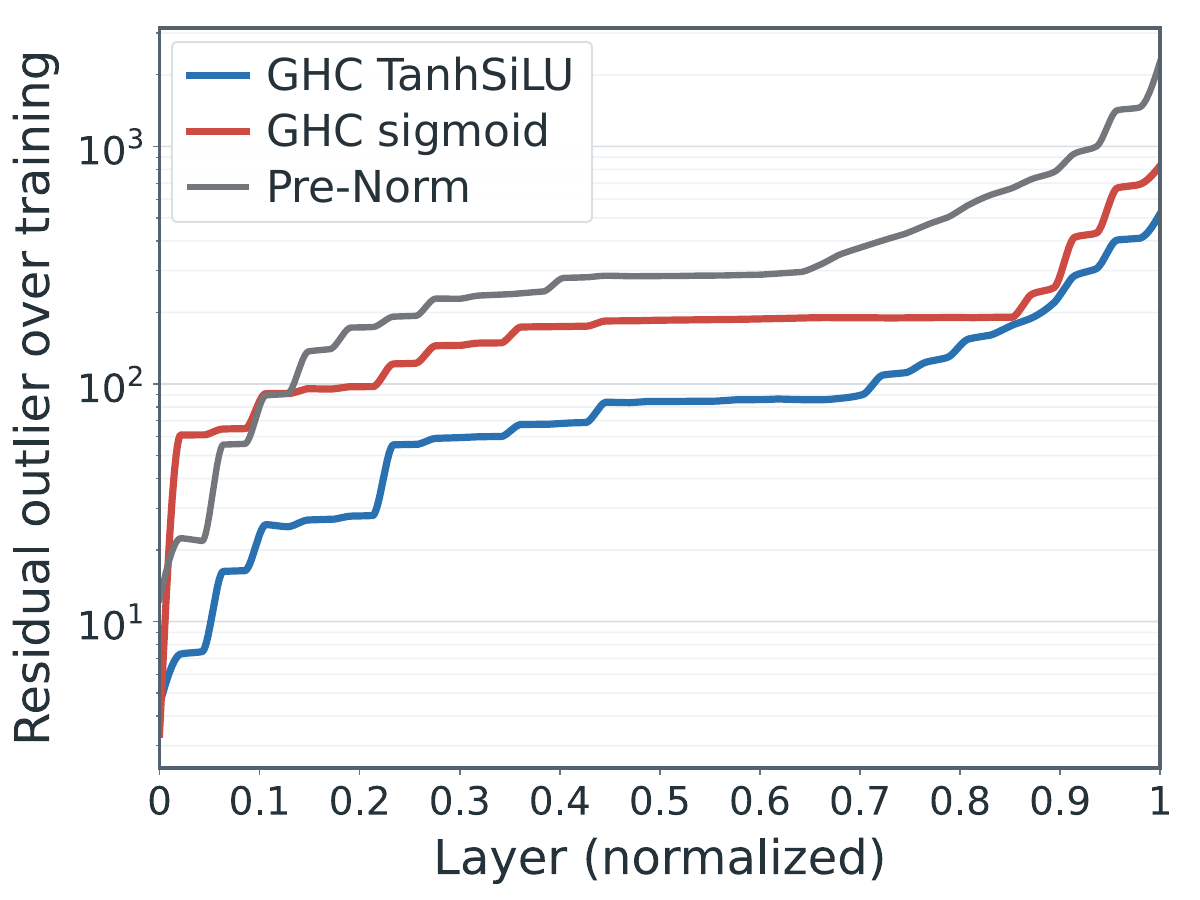}
    \caption{
        Peak residual magnitude ($\max \lvert X_\ell \rvert$ recorded over training) across normalized layer depth on the 430M dense model for Pre-Norm and \method{} with sigmoid or $\operatorname{TanhSiLU}$ gating.
    }
    \vspace{-1.3em}
    \label{fig:residual_outliers}
\end{figure}

%% file: secs/analysis.tex
\section{Analysis}
\label{sec:analysis}

To understand how \method{} utilizes its expanded signed interfaces in practice, we analyze the internal behavior of the trained 430M dense model from two perspectives: how parallel groups specialize their read and write maps across depth (Section~\ref{sec:group_stream_specialization}), and how signed gating reshapes residual magnitude dynamics across layers (Section~\ref{sec:outlier_analysis}).

\subsection{Learned Gating Patterns Across Depth}
\label{sec:group_stream_specialization}

Figure~\ref{fig:group_specialization} visualizes token-averaged read maps $H^{\mathrm{pre}}$ and write maps $H^{\mathrm{post}}$ across representative sublayers (see Appendix~\ref{app:full_gate_maps} for all sublayers).
Since group-wise $\operatorname{RMSNorm}$ is invariant to positive row scaling, we $\ell_2$-normalize each row of $H^{\mathrm{pre}}$ prior to averaging to remove arbitrary scale differences while preserving signs and relative stream weights.
Two empirical patterns emerge:

\noindent\textbf{Signed stream combinations.}
Across trained layers, $H^{\mathrm{pre}}$ develops substantial negative weights (blue) alongside positive weights (red), such as the opposing signs across streams in Layer 5 (Attn) and Layer 12 (FFN).
Rather than restricting groups to nonnegative linear combinations, the model actively exploits the negative range of $\operatorname{TanhSiLU}$ to extract subtractive, differential features across streams.

\noindent\textbf{Group differentiation in deeper layers.}
In Layer 0 (FFN), where all streams have just branched from the shared token embedding, the eight groups exhibit nearly identical read weights while keeping $H^{\mathrm{post}}$ near zero.
As depth increases (Layers 12, 19, and 23), rows within the same sublayer diverge markedly from both one another and the block-diagonal initialization.
In Layer 19 (Attn) and Layer 23 (FFN), parallel groups attend to distinct signed combinations of streams, while $H^{\mathrm{post}}$ (e.g., Layer 19 Attn) routes group updates back with differentiated per-stream weights instead of updating all streams uniformly.

\subsection{Residual Outlier Magnitude}
\label{sec:outlier_analysis}

Figure~\ref{fig:residual_outliers} tracks the residual outlier magnitude~\citep{gatednorm}, measured by the peak entry $\max \lvert X_\ell \rvert$ recorded over training, across normalized layer depth for Pre-Norm and \method{} with sigmoid or $\operatorname{TanhSiLU}$ gating.

Compared with Pre-Norm, sigmoid-gated \method{} lowers peak magnitudes in deeper layers, yet still exhibits a sharp early surge (exceeding $60$ at normalized depth $<0.1$).
Replacing sigmoid with $\operatorname{TanhSiLU}$ yields a substantially flatter growth profile across all depths.
Two properties of signed gating may help explain this behavior.
First, unlike nonnegative sigmoid gates, negative $\operatorname{TanhSiLU}$ weights permit direct subtractive mixing across streams, providing an explicit mechanism to offset shared components and moderate stream growth.
Second, more expressive stream reading and writing may lessen the need for large activation magnitudes to preserve or route signals across depth~\citep{gatednorm}.

%% file: secs/appendix.tex
\appendix
\onecolumn
\section{Detailed Comparison of Residual Information Flow}
\label{app:architecture_diagram}

\begin{figure*}[!htbp]
    \centering
    \includegraphics[width=\linewidth]{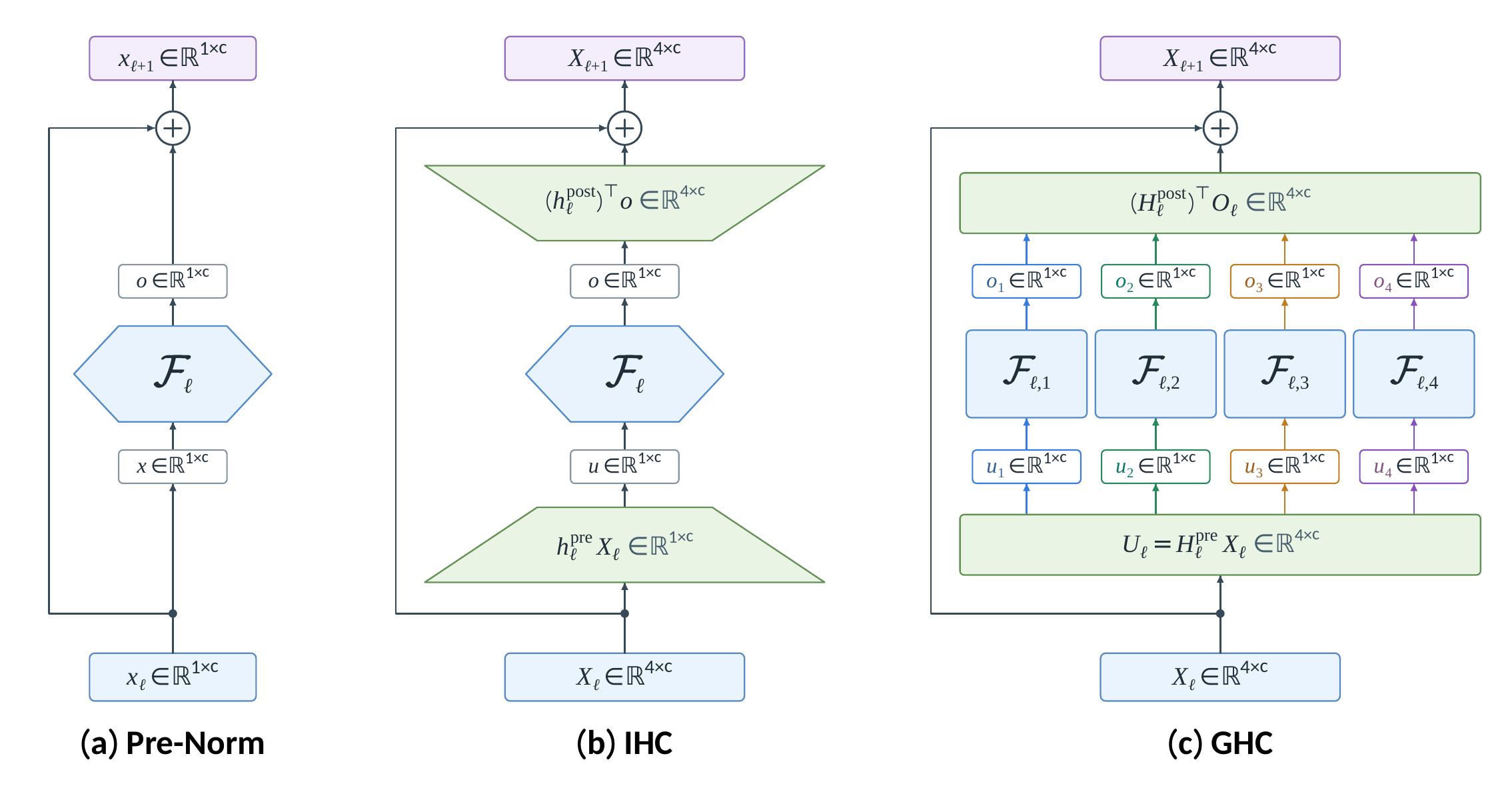}
    \caption{
        Residual interfaces in Pre-Norm, IHC, and \method{} with $n=r=4$.
        Left: Pre-Norm maintains a single $c$-dimensional residual stream and computes one branch output.
        Middle: IHC expands the residual state to four streams, reads one shared mixture, and writes scaled copies of one branch output back to the streams.
        Right: \method{} reads four group-specific mixtures, partitions the same total intermediate width $(M_1,M_2)$ among four computation groups, and distributes their separate outputs back to the streams with learned weights.
        Each group applies input RMSNorm (sharing parameters across groups) and its own slice of the branch projections, with intermediate width $(M_1/4,M_2/4)$ and output width $c$.
        All three architectures use identity stream shortcuts.
    }
    \label{fig:interface}
\end{figure*}

Figure~\ref{fig:interface} provides a detailed schematic comparing the per-layer residual information flow and tensor shapes across Pre-Norm, IHC, and \method{} ($n=r=4$).

\section{A Unified View of Residual Aggregation}
\label{sec:unified_aggregation}

We can express the input to any residual branch as a linear combination of earlier branch outputs~\citep{attentionresiduals}.
As in Section~\ref{sec:Preliminaries}, let $\mathcal{F}_\ell$ denote residual branch $\ell$ (including its input normalization), mapping a per-token input $u_\ell \in \mathbb{R}^{1 \times c}$ to an output $o_\ell = \mathcal{F}_\ell(u_\ell) \in \mathbb{R}^{1 \times c}$, and set $o_0 = e$ for the initial token embedding.
For a given token, standard Pre-Norm, HC, and Attention Residuals all admit the unified form
\begin{equation}
    u_\ell
    = \sum_{i=0}^{\ell-1} \alpha_{\ell,i}\, o_i,
    \label{eq:unified_aggregation}
\end{equation}
where $\alpha_{\ell,i} \in \mathbb{R}$ is the effective scalar weight with which branch $\ell$ reads $o_i$.
These architectures differ in how $\alpha_{\ell,i}$ is parameterized and which intermediate states are maintained across depth.

\noindent\textbf{Standard residual connections.}
Unrolling the Pre-Norm update in Eq.~\eqref{eq:standard_residual} from $x_1 = e$ gives
\begin{equation}
    u_\ell = x_\ell
    = e + \sum_{i=1}^{\ell-1} o_i.
    \label{eq:residual_output_view}
\end{equation}
Thus $\alpha_{\ell,i} = 1$ for all $0 \le i < \ell$, and a single running sum suffices to form every branch input.

\noindent\textbf{Hyper-Connections.}
HC~\citep{hyperconnections} maintains $n$ parallel streams $X_\ell \in \mathbb{R}^{n \times c}$.
To track the contribution of $o_i$ across subsequent stream transitions, define the ordered matrix product
\begin{equation}
    T_{b:a}
    = H_b^{\mathrm{res}}\cdots H_a^{\mathrm{res}},
    \qquad T_{b:b+1} = I_n,
    \label{eq:transition_product}
\end{equation}
where for $b \ge a$ the rightmost factor acts first.
Starting from $X_1 = \mathbf{1}_n e$, unrolling Eq.~\eqref{eq:hc} yields the residual state entering branch $\ell$:
\begin{equation}
    X_\ell
    = T_{\ell-1:1}\mathbf{1}_n e
    + \sum_{i=1}^{\ell-1}
      T_{\ell-1:i+1}(h_i^{\mathrm{post}})^\top o_i.
    \label{eq:hc_unrolled_state}
\end{equation}
Applying the read vector $h_\ell^{\mathrm{pre}} \in \mathbb{R}^{1 \times n}$ recovers Eq.~\eqref{eq:unified_aggregation} with effective scalar weights
\begin{equation}
    \begin{aligned}
        \alpha_{\ell,0}
        &= h_\ell^{\mathrm{pre}} T_{\ell-1:1}\mathbf{1}_n, \\
        \alpha_{\ell,i}
        &= h_\ell^{\mathrm{pre}} T_{\ell-1:i+1}
           (h_i^{\mathrm{post}})^\top,
           \quad 1 \le i < \ell.
    \end{aligned}
    \label{eq:hc_effective_weights}
\end{equation}
Thus the effective weight on $o_i$ factors through how branch $i$ writes to the streams, how intervening layers mix them, and how branch $\ell$ reads them.
In IHC ($H_j^{\mathrm{res}} = I_n$), this simplifies to $\alpha_{\ell,0} = h_\ell^{\mathrm{pre}}\mathbf{1}_n$ and $\alpha_{\ell,i} = h_\ell^{\mathrm{pre}}(h_i^{\mathrm{post}})^\top$ for $1 \le i < \ell$.
Under dynamic gating, these expressions hold with the token-dependent vectors evaluated along the forward pass.

\noindent\textbf{Attention Residuals.}
Full Attention Residuals (Full AttnRes)~\citep{attentionresiduals} explicitly caches the token embedding and all preceding branch outputs.
Branch $\ell$ scores each cached vector using a learned query $w_\ell \in \mathbb{R}^{1 \times c}$ and normalizes the scores via softmax across depth:
\begin{equation}
    \begin{aligned}
        s_{\ell,i}
        &= w_\ell \operatorname{RMSNorm}(o_i)^\top, \\
        \alpha_{\ell,i}
        &= \frac{\exp(s_{\ell,i})}
                {\sum_{j=0}^{\ell-1}\exp(s_{\ell,j})}.
    \end{aligned}
    \label{eq:attnres_weights}
\end{equation}
Block AttnRes reduces memory traffic by attending over block-level partial sums rather than individual branch outputs, which assigns a shared attention weight to all branches within the same block.

\noindent\textbf{Implications for residual interfaces.}
Although HC selects across $n$ streams and Full AttnRes selects across all $\ell$ historical outputs, both still route every branch through a single $c$-dimensional input $u_\ell$ and a single $c$-dimensional output $o_\ell$.
This corresponds directly to the $c$-dimensional interface bottleneck analyzed in Section~\ref{sec:interface_bottlenecks}.
While Eq.~\eqref{eq:unified_aggregation} uses channel-shared scalar weights, channel-wise~\citep{qwen3p8next} and slice-wise~\citep{vwn} variants generalize $\alpha_{\ell,i}$ to diagonal or block-selection operators while still feeding a single backbone-width input into the monolithic branch.

\noindent\textbf{Grouped residual aggregation.}
\method{} lifts this bottleneck by partitioning $\mathcal{F}_\ell$ into $r$ parallel groups $\{\mathcal{F}_{\ell,g}\}_{g=1}^r$ along attention heads or FFN intermediate channels.
Each group $g$ reads its own stream mixture $u_{\ell,g} \in \mathbb{R}^{1 \times c}$ and produces a dedicated output $o_{\ell,g} = \mathcal{F}_{\ell,g}(u_{\ell,g}) \in \mathbb{R}^{1 \times c}$.
Unrolling Eq.~\eqref{eq:expanded} shows that group $g$ of branch $\ell$ aggregates all preceding group outputs with pair-specific weights:
\begin{equation}
    u_{\ell,g}
    = \alpha_{\ell,g;0}\,e
    + \sum_{i=1}^{\ell-1}\sum_{h=1}^{r}
      \alpha_{\ell,g;i,h}\,o_{i,h},
    \label{eq:grouped_unrolled}
\end{equation}
where $\alpha_{\ell,g;0} = (H_\ell^{\mathrm{pre}})_{g,:}\, T_{\ell-1:1}\mathbf{1}_n$ and $\alpha_{\ell,g;i,h} = (H_\ell^{\mathrm{pre}})_{g,:}\, T_{\ell-1:i+1}\, \bigl((H_i^{\mathrm{post}})_{h,:}\bigr)^\top$ for $1 \le i < \ell$ (simplifying to $(H_\ell^{\mathrm{pre}})_{g,:}\bigl((H_i^{\mathrm{post}})_{h,:}\bigr)^\top$ when $H_j^{\mathrm{res}} = I_n$).

Notably, Eq.~\eqref{eq:grouped_unrolled} also suggests a grouped extension of Full AttnRes, in which each query group $g$ attends directly over the embedding and all $r(\ell-1)$ individual group outputs from earlier branches while keeping total intermediate width and branch projection FLOPs unchanged.
However, caching $r$ separate $c$-dimensional vectors per branch and token would multiply the depth-history memory footprint and I/O traffic of Full AttnRes by a factor of $r$ ($rc$ versus $c$ elements per branch and token), making it impractical at scale.

\section{Training Details}
\label{app:training_details}

Our experiments span three settings: a 430M dense comparison, a dense scaling ladder, and two Mixture-of-Experts (MoE) scales (4B and 13B)
In the 430M dense and MoE settings, all methods share identical backbone dimensions, token budgets, global batch sizes, and learning-rate schedules, with peak learning rates and batch sizes determined via StepLaw on the Pre-Norm reference \citep{steplaw}.
For the dense scaling ladder, we apply the StepLaw fitting rule individually to each method and model size to account for connection parameters.

Throughout this section, $N$ denotes the non-embedding parameter count (backbone and connection parameters, excluding token embeddings, the output head, and final normalization), and $A$ denotes the per-token active non-embedding parameter count in MoE models.
Unless noted otherwise, IHC and \method{} use $n=4$ streams, and \method{} uses $r=8$ groups in all attention and dense FFN branches.

\noindent\textbf{Shared architecture and optimizer.}
We train decoder-only language models on packed sequences of length $s=8192$ with zero dropout and a weight initialization standard deviation of $0.006$.
All models use rotary position embeddings~\citep{rope}, elementwise sigmoid attention gates~\citep{gatedattention}, and query/key/value normalization, with head dimension $128$ and $8$ key/value heads.
Layers follow a repeating $3{:}1$ sliding-to-full attention pattern with a $512$-token sliding window ($14$- and $18$-layer models append one additional sliding/full pair).
Dense FFNs use intermediate width $4c$ (for hidden dimension $c$) with SiTU-GLU~\citep{kimik3} activation ($\beta=4$, linear-branch saturation $25$).

Matrix parameters in the backbone are optimized with Muon~\citep{muonisscalable} (Nesterov momentum $0.95$, $8$ Polar Express Newton-Schulz iterations~\citep{polarexpress}) and decoupled weight decay, while all remaining parameters use AdamW ~\citep{adamw} ($\beta_1=0.9, \beta_2=0.95, \epsilon=10^{-15}$).
All runs use weight decay $0.1$, gradient clipping at $1.0$, and a cosine learning-rate schedule with a $2{,}000$-step linear warmup to a floor of $10^{-5}$.

\noindent\textbf{430M dense comparison.}
The 430M backbone ($N=427.88\text{M}$ for Pre-Norm) uses $L=24$ layers, hidden dimension $c=1024$, FFN width $4096$, and $8$ query heads for both sliding and full attention.
All models are trained for $47{,}684$ steps at a global batch size of $128$ sequences ($50\text{B}$ tokens) and a peak learning rate of $2.9\times 10^{-3}$.

\noindent\textbf{Dense scaling ladder.}
We evaluate Pre-Norm, IHC, and \method{} ($r=8$) across depths $L\in\{12,14,16,18\}$ with hidden dimension $c=128L$, FFN width $4c$, and $16$ / $8$ query heads for sliding / full attention.
Each model is trained for $D=20N$ tokens.
Under StepLaw~\citep{steplaw}, the target token batch size $\widehat{B}$ and reference peak learning rate $\widehat{\eta}$ follow
\begin{equation}
    \widehat{B}=0.58\,D^{0.571},
    \qquad
    \widehat{\eta}=1.79\,N^{-0.713}D^{0.307}.
\end{equation}
Rounding the sequence batch size $G$ to the nearest multiple of $128$ gives $G=128$ across all rungs; we rescale the peak learning rate to $\eta_{\mathrm{peak}}=\widehat{\eta}\sqrt{Gs/\widehat{B}}$ and train for $\lceil D/(Gs)\rceil$ steps (Table~\ref{tab:dense_ladder_training}).

\begin{table}[t]
    \centering
    \small
    \setlength{\tabcolsep}{6.5pt}
    \caption{Dense scaling ladder configurations and final validation losses with $c=128L$, FFN width $4c$, global batch size of 1M tokens, and token budget $D=20N$~\citep{chinchilla}. Parameter counts $N$ are in millions, token budgets $D$ in billions.}
    \label{tab:dense_ladder_training}
    \begin{tabular}{@{}lcrrcrc@{}}
        \toprule
        Method & $L$ & $N$ (M) & $D$ (B) & $\eta_{\mathrm{peak}}$ & Steps & Avg.\ Loss $\downarrow$ \\
        \midrule
        \multirow{4}{*}{Pre-Norm}
        & 12 &  476.6 &  9.53 & 2.56$\times 10^{-3}$ &  9,091 & 2.2437 \\
        & 14 &  723.0 & 14.46 & 1.92$\times 10^{-3}$ & 13,792 & 2.1378 \\
        & 16 & 1048.6 & 20.97 & 1.48$\times 10^{-3}$ & 20,002 & 2.0522 \\
        & 18 & 1451.1 & 29.02 & 1.19$\times 10^{-3}$ & 27,677 & 1.9796 \\
        \midrule
        \multirow{4}{*}{IHC}
        & 12 &  477.9 &  9.56 & 2.56$\times 10^{-3}$ &  9,117 & 2.2122 \\
        & 14 &  724.9 & 14.50 & 1.92$\times 10^{-3}$ & 13,826 & 2.1113 \\
        & 16 & 1051.0 & 21.02 & 1.48$\times 10^{-3}$ & 20,047 & 2.0266 \\
        & 18 & 1454.0 & 29.08 & 1.18$\times 10^{-3}$ & 27,734 & 1.9546 \\
        \midrule
        \multirow{4}{*}{\method{}}
        & 12 &  486.2 &  9.72 & 2.53$\times 10^{-3}$ &  9,274 & \textbf{2.1964} \\
        & 14 &  736.1 & 14.72 & 1.90$\times 10^{-3}$ & 14,040 & \textbf{2.0910} \\
        & 16 & 1065.7 & 21.31 & 1.47$\times 10^{-3}$ & 20,327 & \textbf{2.0134} \\
        & 18 & 1472.6 & 29.45 & 1.17$\times 10^{-3}$ & 28,089 & \textbf{1.9415} \\
        \bottomrule
    \end{tabular}
\end{table}
\noindent\textbf{MoE experiments.}
We evaluate two MoE scales corresponding to the 4B and 13B models in Table~\ref{tab:main_results}: $L=12, c=1024$ and $L=16, c=1536$.
In both models, the first two layers use dense FFNs of width $4c$, and all remaining layers use $256$ routed experts with top-$8$ routing, expert intermediate width $c/2$, and no shared experts.
Routing uses FP32 logits with normalized sigmoid scores, DeepSeek-style bias updates for load balancing (rate $10^{-3}$), and an expert-parallel auxiliary loss (weight $10^{-3}$).
Table~\ref{tab:moe_training} lists full specifications.

\begin{table}[t]
    \centering
    \small
    \setlength{\tabcolsep}{7pt}
    \caption{Architectural specifications and training schedules for the 4B and 13B MoE models.}
    \label{tab:moe_training}
    \begin{tabular}{@{}lrr@{}}
        \toprule
        Specification & 4B MoE & 13B MoE \\
        \midrule
        Total / MoE layers        & 12 / 10 & 16 / 14 \\
        Hidden dimension $c$      & 1024    & 1536    \\
        Dense FFN / Expert width  & 4096 / 512 & 6144 / 768 \\
        Routed / Active experts   & 256 / 8 & 256 / 8 \\
        Sliding / Full Q-heads    & 8 / 8   & 16 / 8  \\
        \midrule
        Non-embedding params $N$  & 4.12B   & 12.93B  \\
        Active params $A$         & 216.6M  & 641.0M  \\
        Training tokens $D$       & 65B  & 192B \\
        Global batch size         & 2.1M   & 4.2M   \\
        Peak learning rate        & $1.4\times 10^{-3}$ & $9\times 10^{-4}$ \\
        Training steps            & 30,980  & 45,848  \\
        \bottomrule
    \end{tabular}
\end{table}

\section{Details on Dynamic Gate Calibration}
\label{app:gate_calibration}

\subsection{Comparison of Gate Parameterizations}
\label{app:gate_comparison}

Table~\ref{tab:gate_parameterizations} summarizes the dynamic read and write gate parameterizations across multi-stream residual architectures and compares them with \method{}.
Existing designs exhibit distinct trade-offs across four key properties:

\noindent\textbf{Boundedness.}
Unbounded parameterizations (HC~\citep{hyperconnections} and VWN~\citep{vwn}, which apply learnable scales and biases outside $\tanh$) allow gate magnitudes to drift unconstrained during training.
Placing parameters inside a bounded outer nonlinearity ($\sigma$ or \textnormal{TanhSiLU}) restricts the maximum magnitude of each mixing coefficient.

\noindent\textbf{Signed mixing.}
Sigmoid-based gates in mHC~\citep{mhc}, Gated Residual (GR)~\citep{qwen3p8next}, and IHC~\citep{hy4preview} restrict mixing coefficients to be strictly positive, confining reads and writes to conical combinations.
Using $\tanh$ or \textnormal{TanhSiLU} enables signed coefficients and subtractive stream interactions.

\noindent\textbf{Width calibration.}
For broadcast scalar gates without width scaling (HC, mHC, and IHC), predictor preactivation updates grow with hidden width $c$ (Section~\ref{app:gate-width}).
In GR, the read gate produces a $c$-dimensional channel-wise vector without cross-channel broadcasting and thus avoids explicit $1/\sqrt{c}$ scaling, whereas its scalar write gate is broadcast across all $c$ channels while scaling only by $1/n$, leaving the write path uncalibrated.
Incorporating $1/\sqrt{d_b}$ in VWN (for slice width $d_b$) or $1/\sqrt{c}$ in \method{} ensures width-invariant $\Theta(1)$ updates on both paths.

\noindent\textbf{Lightweight projection.}
Whereas GR employs a two-layer bottleneck MLP to generate channel-wise read gates (while retaining a single linear projection on its write path), lightweight designs compute all read and write coefficients via single linear projections of the normalized residual state.

By combining \textnormal{TanhSiLU} with explicit $1/\sqrt{c}$ width scaling, \method{} simultaneously achieves bounded outputs, signed mixing, width-calibrated updates, and lightweight computation across both interfaces.
\begin{table*}[t]
    \centering
    \caption{
        Comparison of dynamic read and write gate parameterizations across multi-stream residual architectures.
        Each expression denotes a single scalar mixing coefficient, except for the GR read gate, which outputs a $c$-dimensional channel-wise vector per stream (including its $1/n$ factor) via bottleneck projections $W_d$ and $W_u$.
        Here $z$ denotes the normalized residual input to the predictor ($z \in \mathbb{R}^{d_b}$ for slice width $d_b$ in VWN, and $z \in \mathbb{R}^{nc}$ flattened across all $n$ streams otherwise), $\sigma$ denotes the sigmoid function, $w$ denotes an entry-specific linear projection vector, $\alpha$ denotes a predictor-shared learnable scalar scale, and $s, b$ denote entry-specific learnable scales and biases.
        Note that mHC and IHC share the same read/write gate form and differ only in $H^{\mathrm{res}}$; slashed entries ($\cdot/\cdot$) indicate $H^{\mathrm{pre}}/H^{\mathrm{post}}$ when the two paths differ.
    }
    \label{tab:gate_parameterizations}
    \footnotesize
    \setlength{\tabcolsep}{4pt}
    \renewcommand{\arraystretch}{1.25}
    \begin{tabular*}{\textwidth}{@{\extracolsep{\fill}}lllcccc@{}}
        \toprule
        Method
        & $H^{\mathrm{pre}}$
        & $H^{\mathrm{post}}$
        & Bounded
        & Signed
        & Width-Calib.
        & Lightweight \\
        \midrule

        HC
        & $\alpha\tanh(w^{\top} z)+b$
        & $\alpha\tanh(w^{\top} z)+b$
        & \xmark & \cmark & \xmark & \cmark \\

        mHC
        & $\sigma(\alpha w^{\top} z+b)$
        & $2\sigma(\alpha w^{\top} z+b)$
        & \cmark & \xmark & \xmark & \cmark \\

        VWN
        & $s\tanh(w^{\top} z/\sqrt{d_b})+b$
        & $s\tanh(w^{\top} z/\sqrt{d_b})+b$
        & \xmark & \cmark & \cmark & \cmark \\

        GR
        & $\frac{1}{n}\sigma\!\big(W_u\operatorname{\textnormal{SiLU}}(W_d z/n)\big)$
        & $2\sigma(w^{\top} z/n)$
        & \cmark & \xmark & \cmark/\xmark & \xmark/\cmark \\

        IHC
        & $\sigma(\alpha w^{\top} z+b)$
        & $2\sigma(\alpha w^{\top} z+b)$
        & \cmark & \xmark & \xmark & \cmark \\

        \midrule
        \textbf{\method{} (ours)}
        & $\operatorname{\textnormal{TanhSiLU}}(s w^{\top} z/\sqrt{c}+b)$
        & $\operatorname{\textnormal{TanhSiLU}}(s w^{\top} z/\sqrt{c}+b)$
        & \cmark & \cmark & \cmark & \cmark \\

        \bottomrule
    \end{tabular*}
\end{table*}
\subsection{Width Calibration of Read and Write Gates}
\label{app:gate-width}

\noindent\textbf{Setup.}
Fix a layer $\ell$. Let $n$ denote the number of residual streams,
$r$ the number of computational groups, and $c$ the channel width.
The residual state is $X_\ell\in\mathbb{R}^{n\times c}$.
Let $\operatorname{vec}$ denote row-wise vectorization and define the
predictor input as
$z_\ell=\operatorname{RMSNorm}(\operatorname{vec}(X_\ell))
\in\mathbb{R}^{nc}$, where $\operatorname{RMSNorm}$ denotes
root-mean-square normalization.

The read and write gates are
$H_\ell^{\mathrm{pre}},H_\ell^{\mathrm{post}}
\in\mathbb{R}^{r\times n}$.
Let $U_\ell\in\mathbb{R}^{r\times c}$ contain the group inputs before
branch normalization, and let $O_\ell\in\mathbb{R}^{r\times c}$ stack
the group outputs. Row $i$ of $O_\ell$ is obtained by applying group
$i$'s branch, including its input normalization, to row $i$ of $U_\ell$.
With a residual mixing matrix
$H_\ell^{\mathrm{res}}\in\mathbb{R}^{n\times n}$, the next residual
state $X_{\ell+1}\in\mathbb{R}^{n\times c}$ satisfies
\begin{equation}
  U_\ell=H_\ell^{\mathrm{pre}}X_\ell,
  \qquad
  X_{\ell+1}=H_\ell^{\mathrm{res}}X_\ell
  +(H_\ell^{\mathrm{post}})^\top O_\ell.
  \label{eq:gate-read-write}
\end{equation}

For $q\in\{\mathrm{pre},\mathrm{post}\}$, let
$W_\ell^q\in\mathbb{R}^{rn\times nc}$,
$b_\ell^q\in\mathbb{R}^{rn}$, and $\alpha_\ell^q\in\mathbb{R}$
be the predictor weight, bias, and scalar amplitude, respectively.
The two predictors have separate parameters.
Let $\gamma_q\in\mathbb{R}$ be a width exponent to be determined,
and let $\phi:\mathbb{R}\to\mathbb{R}$ be continuously differentiable
and applied elementwise. Define the preactivation
$t_\ell^q\in\mathbb{R}^{rn}$ by
\begin{equation}
  t_\ell^q=b_\ell^q+\alpha_\ell^q c^{-\gamma_q}W_\ell^qz_\ell,
  \qquad
  \operatorname{vec}(H_\ell^q)=\phi(t_\ell^q).
  \label{eq:gate-predictor}
\end{equation}
Let $L$ be a differentiable scalar loss whose dependence on these
predictors is through the computation above. Define
$D_\ell^{\mathrm{pre}}=\nabla_{U_\ell}L
\in\mathbb{R}^{r\times c}$,
$D_\ell^{\mathrm{post}}=\nabla_{X_{\ell+1}}L
\in\mathbb{R}^{n\times c}$, and
$G_\ell^q=\nabla_{t_\ell^q}L\in\mathbb{R}^{rn}$.
We analyze a single invocation and a plain SGD update of one predictor,
holding $X_\ell$, $z_\ell$, and all other parameters fixed.
With shared weights across tokens, this is the gradient contribution
of one invocation.

\begin{assumption}[Local scales and nondegeneracy]
\label{ass:gate-scales}
Keep $n$ and $r$ fixed as $c\to\infty$.
Write $\|\cdot\|_F$ for the Frobenius norm,
$\|\cdot\|_2$ for the Euclidean norm, and
$\|\cdot\|_\infty$ for the maximum absolute coordinate.
There exist positive constants
$C_A,C_D,C_\phi,\kappa_-,\kappa_+,\alpha_-,\alpha_+$, and $g_-$,
independent of $c$, such that, for all sufficiently large $c$
and both $q\in\{\mathrm{pre},\mathrm{post}\}$,
\begin{align}
  \max\{\|X_\ell\|_F^2,\|O_\ell\|_F^2\}
    &\le C_Ac,
  &\kappa_-c\le\|z_\ell\|_2^2&\le\kappa_+c,
  \label{eq:gate-forward-scales}\\
  \|D_\ell^q\|_F&\le C_Dc^{-1/2},
  &\|\phi'(t_\ell^q)\|_\infty&\le C_\phi,
  \label{eq:gate-backward-scales}\\
  \alpha_-\le|\alpha_\ell^q|&\le\alpha_+,
  &\|G_\ell^q\|_2&\ge g_-.
  \label{eq:gate-nondegeneracy}
\end{align}
The output-energy bound for $O_\ell$ is required separately from that
for $X_\ell$. The backward bounds and gradient lower bounds are
assumptions for each gate; forward normalization alone does not imply them.
\end{assumption}

The backward scale is motivated by the maximal-update parametrization
($\mu$P): Tensor Programs IV studies empirical moments of width-scaled
feature gradients
\citep[Section~6 and Theorem~7.4 of the extended version]{yang2022feature},
while Tensor Programs V gives the corresponding output-layer scaling
\citep[Table~3 and Appendix~J.2.1]{yangmupV}.
Under their applicable conditions, at a fixed training step with fixed
output dimension and bounded loss derivatives, a coordinate root-mean-square
gradient of order $O_p(c^{-1})$ yields a Frobenius norm of order
$O_p(c^{-1/2})$ for a fixed number of groups; $O_p$ denotes boundedness
in probability after rescaling. We impose the deterministic local bound
in \eqref{eq:gate-backward-scales}, rather than claim that these works
establish it for GHC.

\begin{theorem}[A common width exponent for both gates]
\label{thm:gate-width}
Under Assumption~\ref{ass:gate-scales}, update $W_\ell^q$ by plain SGD
with a width-independent learning rate $\eta_W^q>0$.
Let $\Delta_{W^q}t_\ell^q$ denote the resulting preactivation change
on the same input. Then, for each $q\in\{\mathrm{pre},\mathrm{post}\}$,
\begin{equation}
  \|\Delta_{W^q}t_\ell^q\|_2
  =\Theta(c^{1-2\gamma_q}),
  \label{eq:gate-width-result}
\end{equation}
where $\Theta$ denotes positive upper and lower bounds independent of
width after rescaling. Consequently, both preactivation updates remain
bounded and bounded away from zero if and only if
$\gamma_{\mathrm{pre}}=\gamma_{\mathrm{post}}=1/2$.
Thus both predictors use the factor $1/\sqrt{c}$ under this criterion.
\end{theorem}

\begin{proof}
The matrix chain rule applied to \eqref{eq:gate-read-write} gives
\[
  \nabla_{H_\ell^{\mathrm{pre}}}L
  =D_\ell^{\mathrm{pre}}X_\ell^\top,
  \qquad
  \nabla_{H_\ell^{\mathrm{post}}}L
  =O_\ell(D_\ell^{\mathrm{post}})^\top.
\]
Here $D_\ell^{\mathrm{pre}}$ includes backpropagation through the
grouped branches and the write operation.
For either gate, the Frobenius product bound implies
\[
  \|\nabla_{H_\ell^q}L\|_F
  \le (C_Dc^{-1/2})(\sqrt{C_A}\,c^{1/2})
  =C_D\sqrt{C_A}.
\]
Let $\odot$ denote elementwise multiplication.
Since vectorization preserves the Frobenius norm,
\eqref{eq:gate-predictor} and the chain rule yield
\begin{align*}
  G_\ell^q
    &=\phi'(t_\ell^q)\odot
      \operatorname{vec}(\nabla_{H_\ell^q}L),\\
  g_-\le\|G_\ell^q\|_2
    &\le\|\phi'(t_\ell^q)\|_\infty
      \|\nabla_{H_\ell^q}L\|_F
    \le C_\phi C_D\sqrt{C_A}.
\end{align*}

Fix either $q$. Differentiating the affine predictor in
\eqref{eq:gate-predictor} gives the outer-product gradient
\[
  \nabla_{W_\ell^q}L
  =\alpha_\ell^q c^{-\gamma_q}G_\ell^qz_\ell^\top.
\]
Substituting the SGD weight change
$\Delta W_\ell^q=-\eta_W^q\nabla_{W_\ell^q}L$ gives
\begin{align}
  \Delta_{W^q}t_\ell^q
    &=\alpha_\ell^q c^{-\gamma_q}\Delta W_\ell^qz_\ell\notag\\
    &=-\eta_W^q(\alpha_\ell^q)^2c^{-2\gamma_q}
      G_\ell^qz_\ell^\top z_\ell\notag\\
    &=-\eta_W^q(\alpha_\ell^q)^2c^{-2\gamma_q}
      \|z_\ell\|_2^2G_\ell^q.
  \label{eq:gate-exact-update}
\end{align}
This identity is exact for a finite step because the preactivation
is affine in the updated weight when all other quantities are fixed.
Taking norms and applying the assumed bounds yields
\[
  \eta_W^q\alpha_-^2\kappa_-g_-\,c^{1-2\gamma_q}
  \le\|\Delta_{W^q}t_\ell^q\|_2
  \le\eta_W^q\alpha_+^2\kappa_+C_\phi C_D\sqrt{C_A}
      \,c^{1-2\gamma_q}.
\]
Both coefficients are positive and independent of $c$, proving
\eqref{eq:gate-width-result}. The update vanishes for $\gamma_q>1/2$,
diverges in norm for $\gamma_q<1/2$, and remains $\Theta(1)$ precisely
when $\gamma_q=1/2$. This applies separately to both gates.
\end{proof}

Each mixing coefficient is shared across the $c$ hidden channels,
so its gradient is an inner product over those channels.
At $\gamma_q=1/2$, the weight gradient is proportional to
$\alpha_\ell^q$, whereas the scale gradient is
\[
  \frac{\partial L}{\partial \alpha_\ell^q}
  = \frac{1}{\sqrt{c}}
    \big\langle G_\ell^q, W_\ell^q z_\ell \big\rangle .
\]
Thus a small initial $\alpha_\ell^q$ directly attenuates the
gradient to $W_\ell^q$, while the gradient to $\alpha_\ell^q$
has no such explicit factor.
Moreover, replacing the shared scalar $\alpha_\ell^q$ with entry-wise
scales $s_\ell^q\in\mathbb{R}^{r\times n}$ as in
Eq.~\eqref{eq:calibrated_gate} (flattened as
$\tilde{s}_\ell^q=\operatorname{vec}(s_\ell^q)\in\mathbb{R}^{rn}$)
substitutes $\operatorname{diag}(\tilde{s}_\ell^q)^2$ for
$(\alpha_\ell^q)^2$ in \eqref{eq:gate-exact-update} and decouples the
scale gradient across entries:
$\nabla_{\tilde{s}_\ell^q}L = c^{-\gamma_q} G_\ell^q \odot (W_\ell^q z_\ell)$.
Whenever every coordinate of $\tilde{s}_\ell^q$ satisfies the uniform
bounds in \eqref{eq:gate-nondegeneracy}, the sandwich bound above holds
unchanged and yields the identical exponent $\gamma_q=1/2$, while
avoiding cross-entry gradient accumulation onto a single scalar.

The result concerns local preactivation updates in the stated parameter
coordinates under fixed SGD learning rates. Nondegeneracy must hold for
each gate; the fixed first-layer read gate is excluded.
It does not establish nonvanishing feature changes after branch
normalization or long-term stability of the full network.

\subsection{Gate Activations and Their Gradients}
\label{sec:gate_activation_gradients}

\begin{figure*}[t]
    \centering
    \includegraphics[width=\linewidth]{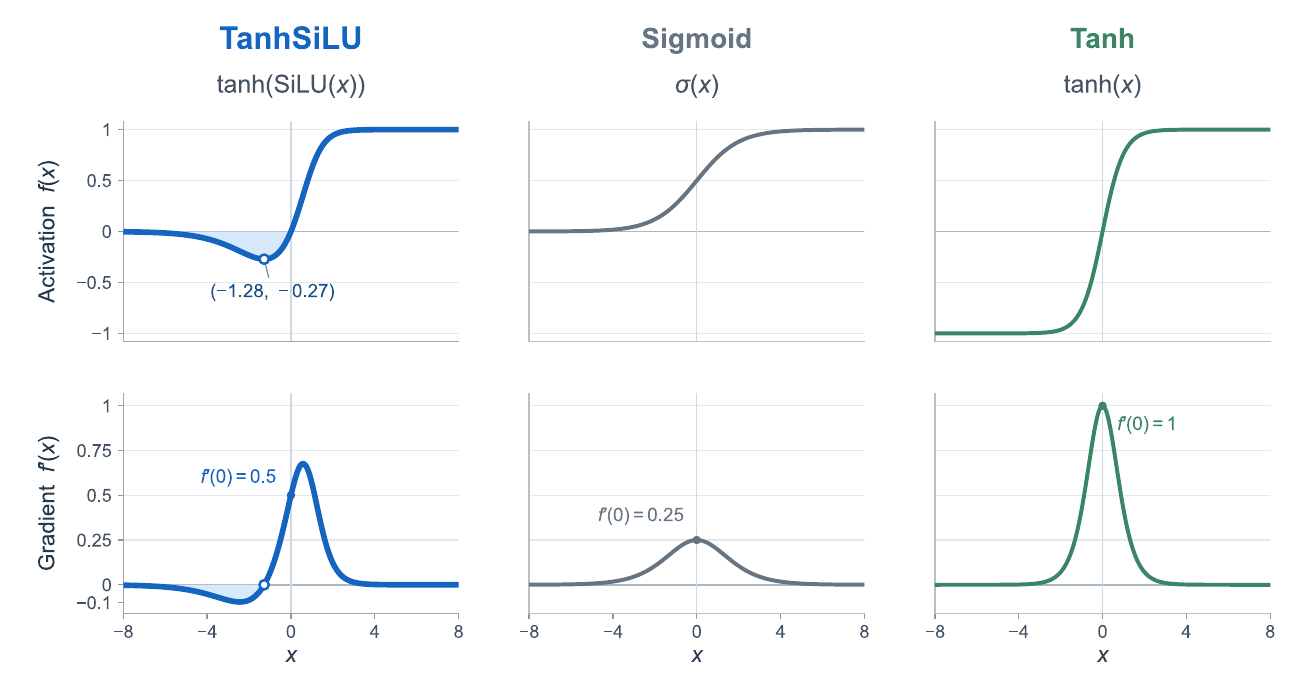}
    \caption{
        Comparison of $\operatorname{TanhSiLU}$, sigmoid and $\tanh$ gate activations (top) and their derivatives with respect to the preactivation $t$ (bottom), plotted on shared axes within each row.
        Filled circles mark values and slopes at the origin ($t=0$), and open circles align the $\operatorname{TanhSiLU}$ minimum ($t\approx -1.28$) with its zero derivative.
    }
    \label{fig:gate_activation_gradients}
\end{figure*}

Figure~\ref{fig:gate_activation_gradients} compares the activation curves and derivatives of sigmoid, $\tanh$, and $\operatorname{TanhSiLU}$, illustrating their gradient behavior near zero.
For a sigmoid gate $a=\sigma(t)$, driving $a$ toward zero requires $t=\log\bigl(a/(1-a)\bigr) \to -\infty$, where the derivative $\sigma'(t)=a(1-a)$ vanishes.
Suppressing a stream therefore pushes the predictor into saturation, attenuating gradients to the preactivation precisely when the stream is muted.

For $\phi(t)=\operatorname{TanhSiLU}(t)$ in Eq.~\eqref{eq:gate_activation}, the chain rule gives
\begin{equation}
    \phi'(t)
    = \bigl(1-\phi(t)^2\bigr)\sigma(t)
      \bigl[1+t\bigl(1-\sigma(t)\bigr)\bigr].
    \label{eq:tanhsilu_grad}
\end{equation}
Because $\phi(0)=0$ with $\phi'(0)=1/2$, a group can set a read or write weight to zero at the origin while retaining responsive first-order gradients in both positive and negative directions.
Note that this non-vanishing sensitivity is local to the zero crossing at $t=0$; along the far negative tail ($t \to -\infty$), $\phi(t)$ also approaches zero with vanishing derivative.

While $\tanh$ shares a finite zero crossing at the origin (with $\tanh(0)=0$ and $\tanh'(0)=1$), it treats positive and negative weights symmetrically over $(-1, 1)$.
In contrast, $\operatorname{TanhSiLU}$ is asymmetric: it attains a bounded minimum of approximately $-0.27$ at $t\approx -1.28$ while approaching $1$ as $t\to\infty$, supporting full-strength additive aggregation alongside bounded subtractive corrections.
\par\vspace{\baselineskip}

\section{Ablation Studies on Mixture-of-Experts}
\label{app:moe_ablation}

To validate our MoE adaptation in Section~\ref{subsec:method}, we evaluate six multi-stream configurations alongside the single-stream Pre-Norm reference on the 4B MoE model (12 attention layers, 2 early dense FFN layers, and 10 MoE layers with $E=256$ routed experts and $k=8$ active experts per token).
Starting from IHC, these variants progressively test naive expert partitioning against global top-$k$ routing, and disentangle wider router inputs from expert-specific residual interfaces:

\noindent\textbf{IHC}: Applies shared dynamic interfaces ($n=4, r=1$) to all branches, with global top-8 routing ($c \to 256$) over the MoE read mixture.

\noindent\textbf{IHC + Group MoE}: Splits the 256 experts into 8 disjoint subsets of 32 with independent top-1 routers ($8 \times \text{top-1/32}$), while keeping shared $r=1$ IHC interfaces.

\noindent\textbf{\method{} + Group MoE}: Extends dense \method{} ($r=8$) directly to MoE by pairing each of the 8 dynamic read/write groups with its own disjoint top-1/32 expert subset.

\noindent\textbf{\method{} (MoE $r=1$)}: Applies \method{} ($r=8$) to attention and dense FFN branches, while keeping MoE layers at $r=1$ with global top-8 routing ($c \to 256$) and shared dynamic mixing.

\noindent\textbf{\method{} (MoE $r=1$) + $4c$ Router}: Expands the global router of \method{} (MoE $r=1$) to condition on all concatenated streams ($4c \to 256$), matching the router parameter count of full \method{} while keeping shared $r=1$ expert interfaces.

\noindent\textbf{\method{} (ours)}: Pairs the stream-concatenated global router ($4c \to 256$) with $2n$ static learnable parameters per expert (passed through $\operatorname{TanhSiLU}$) for expert-specific stream reading and writing.

\begin{table*}[t]
    \centering
    \caption{
        Ablation of MoE routing and residual interface designs on the 4B MoE model ($n=4$, 65B tokens).
        $\Delta_{\mathrm{IHC}}$ denotes validation loss change relative to the global-routing IHC baseline.
    }
    \label{tab:moe_ablation}
    \small
    \setlength{\tabcolsep}{6.5pt}
    \begin{tabular}{@{}llllcc@{}}
        \toprule
        Method & Attn \& Dense  & MoE Routing & MoE Read / Write & Avg.\ Loss $\downarrow$ & $\Delta_{\mathrm{IHC}}$ \\
        \midrule
        \grayit{Pre-Norm}                   & \grayit{Pre-Norm} & \grayit{Global top-8 ($c \to 256$)} & \grayit{Single stream}      & \grayit{1.9260} & \grayit{+0.0194} \\
        IHC                                 & IHC ($r=1$)       & Global top-8 ($c \to 256$)          & Shared dynamic ($r=1$)      & 1.9066          & \phantom{+}0.0000 \\
        IHC + Group MoE                     & IHC ($r=1$)       & Grouped $8 \times \text{top-1/32}$  & Shared dynamic ($r=1$)      & 1.9179          & +0.0113 \\
        \method{} + Group MoE               & \method{} ($r=8$) & Grouped $8 \times \text{top-1/32}$  & Group dynamic ($r=8$)       & 1.9023          & $-$0.0043 \\
        \method{} (MoE $r=1$)               & \method{} ($r=8$) & Global top-8 ($c \to 256$)          & Shared dynamic ($r=1$)      & 1.8945          & $-$0.0121 \\
        \method{} (MoE $r=1$) + $4c$ Router & \method{} ($r=8$) & Global top-8 ($4c \to 256$)         & Shared dynamic ($r=1$)      & 1.8960          & $-$0.0106 \\
        \method{} (ours)                    & \method{} ($r=8$) & Global top-8 ($4c \to 256$)         & Per-expert static           & \textbf{1.8862} & \textbf{$-$0.0204} \\
        \bottomrule
    \end{tabular}
\end{table*}

Table~\ref{tab:moe_ablation} reports the validation losses across these configurations. Comparing the controlled pairs along the rows of Table~\ref{tab:moe_ablation} yields two main observations.

\noindent\textbf{Splitting non-separable top-$k$ routing degrades MoE quality.}
Unlike per-head attention and elementwise FFN activations, top-$k$ expert selection operates globally across all experts.
Forcibly partitioning the 256 experts into $8 \times \text{top-1/32}$ sub-routers alters this nonlinear competition structure whenever multiple relevant experts fall within the same subset, degrading IHC loss by $0.0113$ ($1.9066 \to 1.9179$).
Although equipping the partitioned groups with $r=8$ dynamic interfaces (\method{} + Group MoE) recovers $0.0156$ over IHC + Group MoE ($1.9179 \to 1.9023$), this structural routing penalty caps its net gain over standard IHC at $0.0043$.

\noindent\textbf{Preserving model properties while partitioning only linear interfaces drives the gains.}
\method{} achieves strong gains precisely when non-separable branch nonlinearities are preserved and only linear pathways are decoupled.
Simply retaining intact global top-8 routing while partitioning the linear projections of attention and dense FFNs (\method{}, MoE $r=1$) already surpasses \method{} + Group MoE by $0.0078$ and lowers loss by $0.0121$ over IHC ($1.9066 \to 1.8945$).
Within the MoE layers, preserving global top-8 competition while decoupling each expert's linear stream read and write combinations (\method{}) delivers a further $0.0083$ reduction to $1.8862$ ($-0.0204$ over IHC).
By contrast, widening the linear router input alone while keeping shared $r=1$ expert interfaces (\method{} ,MoE $r=1$ with $4c$ Router, $1.8960$) yields no gain, confirming that decoupling linear expert interfaces while preserving standard routing semantics is essential.

Finally, our use of static rather than dynamic expert weights reflects communication constraints under expert parallelism (EP): keeping cross-device expert transmission bounded requires performing stream linear combinations locally, whereas storing dynamic gate predictors for all $E=256$ experts locally on every EP rank would incur substantial $\mathcal{O}(En^2c)$ parameter and memory-access overhead.
We therefore leave token-dependent expert gating for future exploration.
\section{Evaluation Details}
\label{sec:evaluation_details}

Table~\ref{tab:eval_protocol} summarizes the sample counts, few-shot settings, and evaluation metrics across all benchmarks.

\begin{table}[t]
    \centering
    \caption{Evaluation benchmarks and metrics.}
    \label{tab:eval_protocol}
    \small
    \setlength{\tabcolsep}{10pt}
    \renewcommand{\arraystretch}{1.15}
    \begin{tabular}{@{}l r r l@{}}
        \toprule
        \textbf{Benchmark} & \textbf{Examples} & \textbf{Shots} & \textbf{Metric} \\
        \midrule
        LAMBADA   & 5,153  & 0 & Final-word Perplexity \\
        PIQA      & 1,838  & 0 & Length-norm.\ Accuracy \\
        HellaSwag & 10,042 & 0 & Length-norm.\ Accuracy \\
        TriviaQA  & 17,944 & 5 & Normalized Exact Match \\
        MATH      & 5,000  & 4 & Mathematical Equivalence \\
        GSM8K     & 1,319  & 8 & Flexible Exact Match \\
        HumanEval & 164    & 0 & pass@1 \\
        \bottomrule
    \end{tabular}
\end{table}
\section{Complete Gate Visualizations Across All Sublayers}
\label{app:full_gate_maps}

\begin{figure*}[!htbp]
    \centering
    \includegraphics[width=\linewidth,height=0.92\textheight,keepaspectratio]{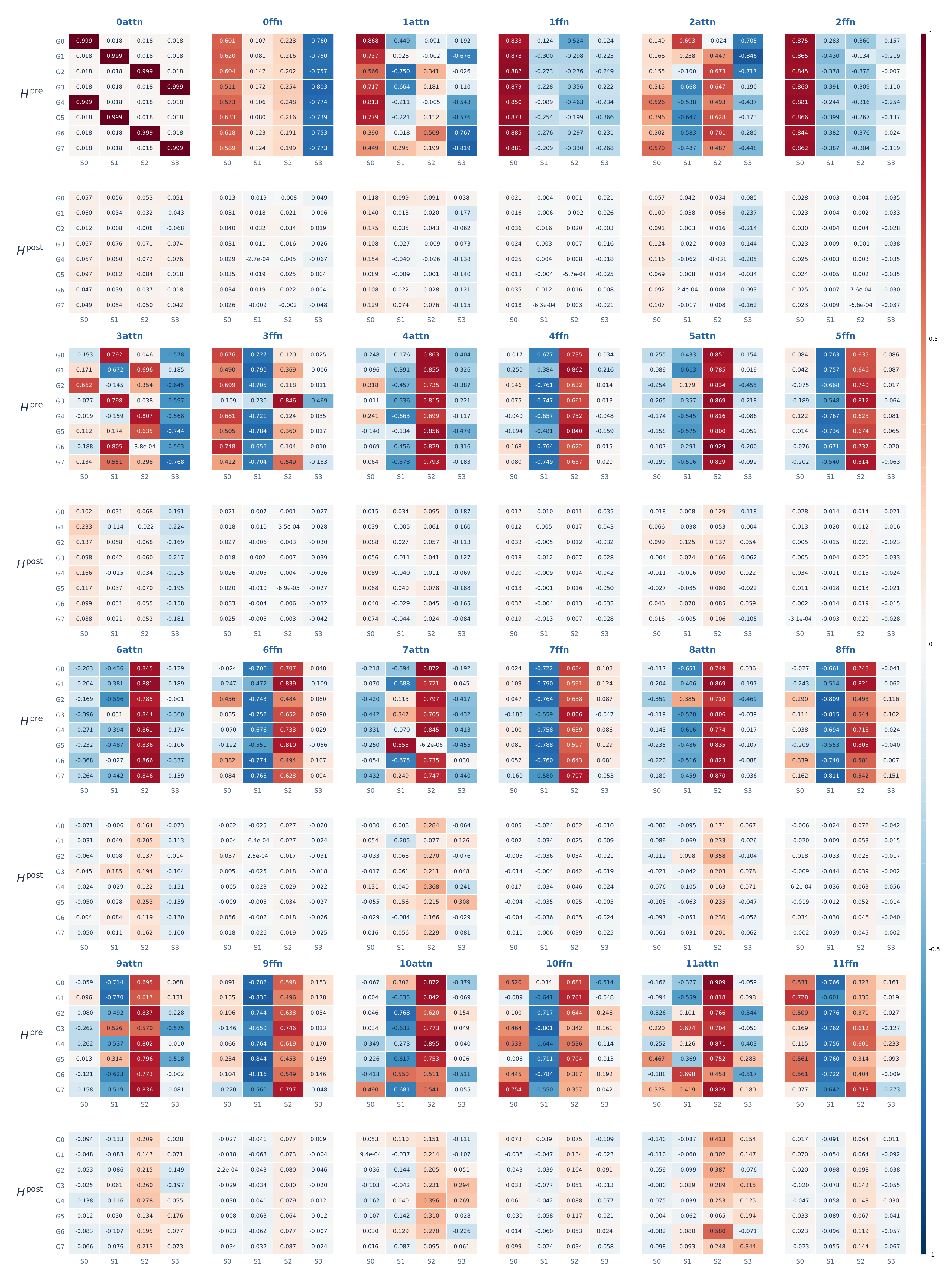}
    \caption{
        Token-averaged read gates $H^{\mathrm{pre}}$ (row-wise $\ell_2$-normalized) and write gates $H^{\mathrm{post}}$ across Layers 0 to 11 ($24$ alternating Attention and FFN sublayers) of the 430M \method{} model ($n=4, r=8, \operatorname{TanhSiLU}$).
        Each $8 \times 4$ heatmap plots the $8$ computation groups (rows) against the $4$ residual streams (columns), with red, white, and blue denoting positive, near-zero, and negative values, respectively.
    }
    \label{fig:full_gates_00_11}
\end{figure*}

\begin{figure*}[!htbp]
    \centering
    \includegraphics[width=\linewidth,height=0.92\textheight,keepaspectratio]{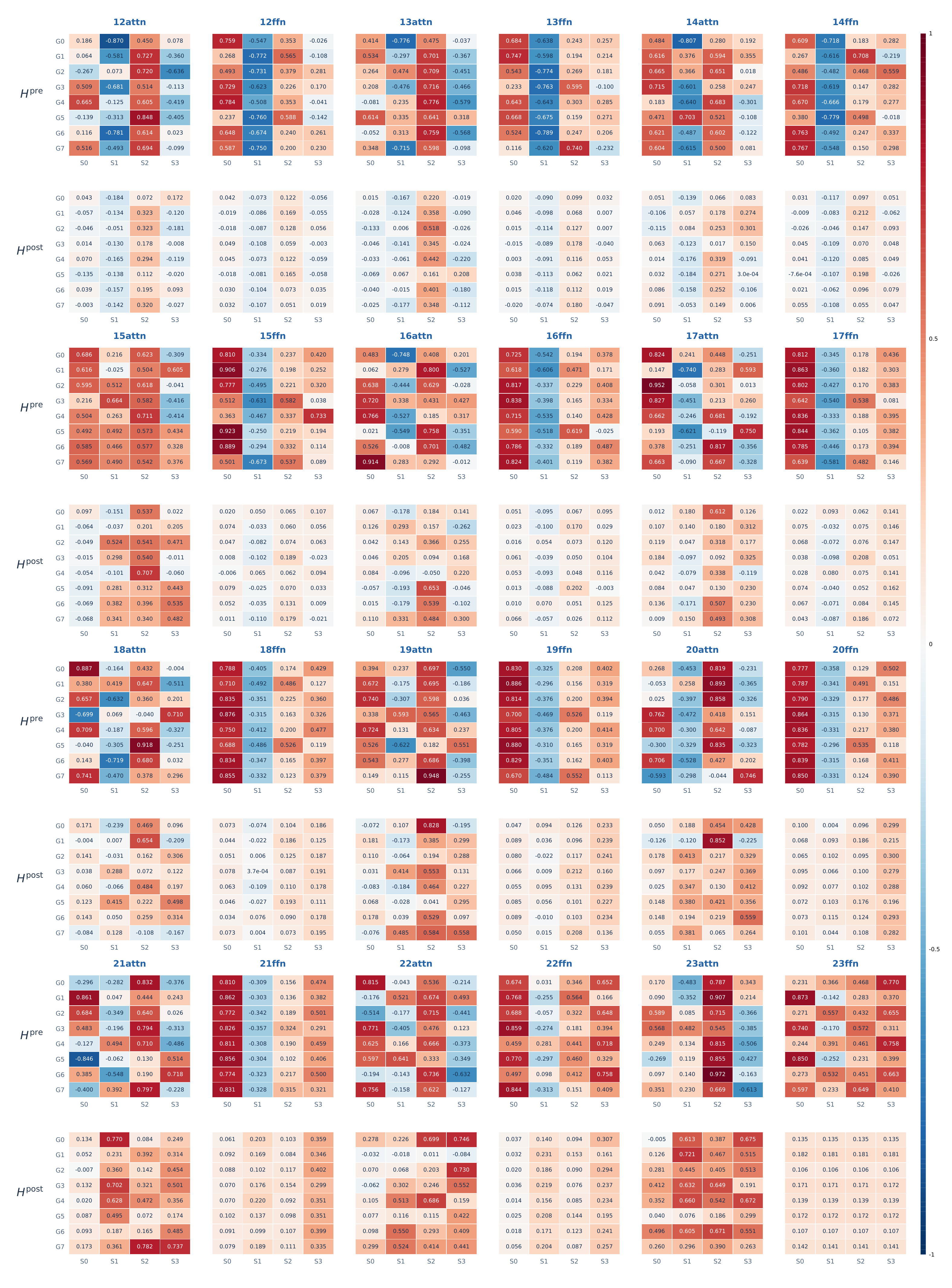}
    \caption{
        Token-averaged read gates $H^{\mathrm{pre}}$ (row-wise $\ell_2$-normalized) and write gates $H^{\mathrm{post}}$ across Layers 12 to 23 ($24$ alternating Attention and FFN sublayers) of the 430M \method{} model ($n=4, r=8, \operatorname{TanhSiLU}$).
        Each $8 \times 4$ heatmap plots the $8$ computation groups (rows) against the $4$ residual streams (columns), with red, white, and blue denoting positive, near-zero, and negative values, respectively.
    }
    \label{fig:full_gates_12_23}
\end{figure*}

Complementing the representative sublayers in Section~6.1 (Figure~\ref{fig:group_specialization}), Figures~\ref{fig:full_gates_00_11} and~\ref{fig:full_gates_12_23} present the token-averaged read maps $H^{\mathrm{pre}}$ and write maps $H^{\mathrm{post}}$ across all 24 layers (48 attention and FFN sublayers in total) of the trained 430M dense \method{} model ($n=4, r=8, \operatorname{TanhSiLU}$), covering Layers 0 to 11 and Layers 12 to 23, respectively.
Consistent with Figure~\ref{fig:group_specialization}, each row $a$ of $H^{\mathrm{pre}}$ is normalized to unit $\ell_2$ norm ($\sum_b a_b^2 = 1$) prior to averaging over validation tokens to factor out positive row-scale invariance under group-wise RMSNorm, whereas $H^{\mathrm{post}}$ is plotted at its raw scale.
Note that the first attention read map (Layer 0 Attn $H^{\mathrm{pre}}$) remains fixed at its initial cyclic pattern during training as described in Section~\ref{sec:gate_calibration}.

\section{Kernel Fusion and Distributed Scaling Benchmarks}
\label{app:kernel_fusion}

To quantify the inference overhead discussed in Section~\ref{sec:infrastructure_efficiency}, we first benchmark custom Triton/CUDA kernels against eager PyTorch on the 430M models ($L=24$, $c=1024$, FFN width $4096$), comparing Pre-Norm, IHC ($n=4,r=1$), and \method{} ($n=4,r=8$).
Input kernels fuse stream mixing, RMS statistics, projections, and FFN SiTU-GLU; output kernels fuse projection, stream mixing, and residual write-back.
All single-GPU end-to-end runs share the same KV-cache policy (18 sliding-window layers of size 512 and six full-attention layers) in BF16 on one H800 GPU, evaluating a $7{,}680$-token prefill and $127$ cached decode steps across $B\in\{2,8,32,128\}$.
As shown in Table~\ref{tab:kernel_fusion_e2e}, kernel fusion accelerates \method{} generation by $2.52\times$ to $2.78\times$ and reduces $B=32$ peak memory from $40.83$ to $17.79\text{ GiB}$ ($56.4\%$).

\begin{table}[t]
    \centering
    \small
    \setlength{\tabcolsep}{4.8pt}
    \caption{Single-GPU end-to-end inference latency and peak allocated memory on the 430M models ($7{,}680$-token prefill and $127$ decode steps on one H800). Speedup is the ratio of total wall-time medians.}
    \label{tab:kernel_fusion_e2e}
    \resizebox{\linewidth}{!}{%
    \begin{tabular}{@{}clrrrrrrrrr@{}}
        \toprule
        & & \multicolumn{2}{c}{Prefill (ms)} & \multicolumn{2}{c}{Decode (ms)} & \multicolumn{2}{c}{Total (ms)} & & \multicolumn{2}{c}{Peak Mem (GiB)} \\
        \cmidrule(lr){3-4} \cmidrule(lr){5-6} \cmidrule(lr){7-8} \cmidrule(l){10-11}
        Batch Size & Method & PyTorch & Fused & PyTorch & Fused & PyTorch & Fused & Speedup & PyTorch & Fused \\
        \midrule
        \multirow{3}{*}{2}
        & Pre-Norm & 129.23 & 63.30 & 1,974.57 & 1,065.92 & 2,103.74 & 1,129.17 & $1.86\times$ & 2.49 & 3.30 \\
        & IHC & 249.76 & 70.20 & 3,101.25 & 1,386.76 & 3,350.79 & 1,456.96 & $2.30\times$ & 2.67 & 3.40 \\
        & \method{} & 387.25 & 83.95 & 3,298.65 & 1,376.93 & 3,686.04 & 1,460.97 & $2.52\times$ & 4.60 & 4.02 \\
        \midrule
        \multirow{3}{*}{8}
        & Pre-Norm & 476.61 & 230.93 & 1,973.65 & 1,064.26 & 2,450.60 & 1,295.01 & $1.89\times$ & 5.86 & 5.82 \\
        & IHC & 947.04 & 266.37 & 3,081.10 & 1,423.12 & 4,027.71 & 1,688.82 & $2.38\times$ & 6.57 & 6.09 \\
        & \method{} & 1,485.84 & 318.75 & 3,237.24 & 1,382.24 & 4,723.84 & 1,701.14 & $2.78\times$ & 11.84 & 6.73 \\
        \midrule
        \multirow{3}{*}{32}
        & Pre-Norm & 1,868.65 & 932.57 & 2,795.53 & 2,187.08 & 4,664.18 & 3,119.83 & $1.50\times$ & 19.36 & 16.10 \\
        & IHC & 3,734.08 & 1,052.38 & 3,342.70 & 2,254.23 & 7,076.63 & 3,307.41 & $2.14\times$ & 22.18 & 17.14 \\
        & \method{} & 5,944.08 & 1,254.73 & 3,387.96 & 2,294.42 & 9,332.55 & 3,549.19 & $2.63\times$ & 40.83 & 17.79 \\
        \midrule
        \multirow{3}{*}{128}
        & Pre-Norm & OOM & 3,753.08 & OOM & 7,778.92 & OOM & 11,532.06 & -- & -- & 57.21 \\
        & IHC & OOM & 4,245.15 & OOM & 7,983.59 & OOM & 12,235.22 & -- & -- & 61.31 \\
        & \method{} & OOM & 5,035.02 & OOM & 8,071.63 & OOM & 13,112.59 & -- & -- & 62.04 \\
        \bottomrule
    \end{tabular}}
\end{table}

\noindent\textbf{Single-GPU end-to-end latency and memory.}
At $B=8$, fused \method{} incurs $+19.7\%$ prefill overhead over IHC with comparable decode latency, yielding nearly identical total generation time ($+0.7\%$ over IHC and $+31.4\%$ over Pre-Norm).
At $B=32$, fused \method{} increases peak memory by only $3.8\%$ over IHC and $10.5\%$ over Pre-Norm, whereas at $B=128$ eager PyTorch runs out of memory across all three architectures.

\begin{table}[t]
    \centering
    \small
    \setlength{\tabcolsep}{5pt}
    \caption{Single-GPU optimized 24-layer projection and residual-interface prefill latency (ms) on one H800 across model scales: 430M ($c=1024$, FFN width $4096$, $B=32, S=7{,}680$, $T=245{,}760$) and Large ($c=4096$, FFN width $16{,}384$, $B=8, S=8{,}192$, $T=65{,}536$). All results incorporate further prefill-specific optimizations. Attention-internal operations (QKV norm, RoPE, SDPA, and output gating) are omitted to isolate projection and stream-interface scaling.}
    \label{tab:prefill_breakdown}
    \label{tab:prefill_scaling}
    \resizebox{\linewidth}{!}{%
    \begin{tabular}{@{}lrrrrrr@{}}
        \toprule
        & \multicolumn{3}{c}{430M ($c=1024$, $T=245{,}760$)} & \multicolumn{3}{c}{Large ($c=4096$, $T=65{,}536$)} \\
        \cmidrule(lr){2-4} \cmidrule(l){5-7}
        Projection / Interface Stage & Pre-Norm & IHC & \method{} & Pre-Norm & IHC & \method{} \\
        \midrule
        HC controller ($2\times$/layer) & 0.00 & 31.49 & 32.40 & 0.00 & 33.31 & 34.06 \\
        Attn input (mix + RMSNorm + QKVG proj.) & 73.46 & 84.07 & 121.82 & 317.86 & 307.51 & 341.14 \\
        Attn output (out proj.\ + write-back) & 28.27 & 48.65 & 68.62 & 83.54 & 98.64 & 159.38 \\
        FFN input (mix + RMSNorm + up/gate proj.\ + SiTU) & 206.50 & 221.04 & 240.55 & 712.01 & 725.74 & 686.62 \\
        FFN output (down proj.\ + write-back) & 76.06 & 95.16 & 144.60 & 305.78 & 337.03 & 366.24 \\
        \midrule
        Projection \& interface total (ms) & 384.29 & 480.41 & 607.99 & 1,419.19 & 1,502.23 & 1,587.44 \\
        Overhead vs.\ Pre-Norm & -- & $+25.0\%$ & $+58.2\%$ & -- & $+5.9\%$ & $+11.9\%$ \\
        \bottomrule
    \end{tabular}}
\end{table}

\noindent\textbf{Single-GPU prefill optimization and width scaling.}
Because prefill accounts for most of the remaining single-GPU overhead, we further optimized the Hopper prefill kernels with single-pass stream mixing and weight-folded RMSNorm, native grouped projection layouts, persistent epilogue write-back where register budgets permit, and CUDA Graph execution, cutting 24-layer projection and interface time on the 430M model ($B=32$) by $16.9\%$ ($731.30\to 607.99\text{ ms}$).
Table~\ref{tab:prefill_scaling} compares all projection and stream-interface stages between 430M ($c=1024$) and a Large 24-layer configuration ($c=4096$, FFN width $16{,}384$) at comparable total hidden volume $Tc$.
As width grows from $c=1024$ to $c=4096$, projection-stage overhead over Pre-Norm drops sharply from $+58.2\%$ to $+11.9\%$ for \method{} and from $+25.0\%$ to $+5.9\%$ for IHC.
This rapid amortization follows directly from arithmetic intensity: dense projections scale quadratically as $\mathcal{O}(Tc^2)$, whereas stream mixing, controller gating, and residual traffic scale linearly as $\mathcal{O}(Tc)$.
On a single GPU, the remaining gap stems from two practical hardware limits: materializing grouped input mixtures in HBM (since fusing mixing into the GEMM prologue reloads the residual across output tiles, while deferring it multiplies prefill FLOPs by $n=4$), and register pressure on wide output reductions alongside $8$-way GEMM dimension splitting.

\begin{table}[t]
    \centering
    \small
    \setlength{\tabcolsep}{6pt}
    \caption{8-GPU distributed 24-layer full-forward prefill latency (ms, including SDPA, RoPE, and attention gating) for the Large dense model ($c=4096$, FFN width $16{,}384$) on an 8$\times$ H800 node (NVLink 4) with two-microbatch compute-communication overlap. Percentages denote overhead relative to Pre-Norm.}
    \label{tab:distributed_prefill}
    \resizebox{\linewidth}{!}{%
    \begin{tabular}{@{}llrrr@{}}
        \toprule
        Parallelism & Workload & Pre-Norm & IHC ($n=4,r=1$) & \method{} ($n=4,r=8$) \\
        \midrule
        \multirow{2}{*}{TP=8 + SP (Attn \& FFN)} & $B=8, S=8{,}192$ & 391.8 & 404.1 ($+3.1\%$) & 403.5 ($+3.0\%$) \\
        & $B=2, S=32{,}768$ & 423.3 & 437.6 ($+3.4\%$) & 436.9 ($+3.2\%$) \\
        \midrule
        \multirow{2}{*}{TP=8 + SP (Attn), Replicated FFN} & $B=8, S=8{,}192$ & 292.4 & 306.9 ($+5.0\%$) & 314.6 ($+7.6\%$) \\
        & $B=2, S=32{,}768$ & 325.4 & 340.4 ($+4.6\%$) & 347.5 ($+6.8\%$) \\
        \bottomrule
    \end{tabular}}
\end{table}

\noindent\textbf{Multi-GPU distributed execution (TP+SP alignment).}
When scaling the Large dense model to 8 GPUs under Tensor Parallelism combined with Sequence Parallelism ($\text{TP}=P=8$ with SP), \method{}'s $r=8$ grouped structure aligns one-to-one with tensor sharding, closing almost all of the remaining efficiency gap.
Table~\ref{tab:distributed_prefill} reports end-to-end 24-layer prefill measurements (including full SDPA) on an 8$\times$ H800 node with two-microbatch compute-communication overlap.
Under standard TP=8 + SP, \method{} reduces overhead over Pre-Norm to just $+3.0\%$ to $+3.2\%$, matching or slightly outperforming IHC across both sequence lengths.
This convergence follows from three complementary mechanisms:

1. \emph{Identical communication volume at $P=r$.} Under SP, each rank stores $T_{\mathrm{loc}} = T/P$ tokens of the $n$-stream residual, computes its $r$ local input mixtures, and dispatches group $g$ to rank $g$ via AllToAll (and symmetrically returns group outputs via AllToAll). When $P=r$, each rank sends $\frac{P-1}{P}Tc$ elements per collective, matching the exact $1.00\times$ per-rank byte volume of Pre-Norm and IHC's AllGather and ReduceScatter.

2. \emph{Zero-SM Copy Engine DMA and elimination of GEMM splitting.} Because each rank now owns exactly one branch, every GPU executes a single monolithic GEMM of the exact same shape as Pre-Norm and IHC, removing the single-GPU grouped-GEMM utilization penalty. Moreover, unlike ReduceScatter, which requires arithmetic reduction and SM involvement in standard collectives, \method{}'s bidirectional AllToAll consists purely of peer-to-peer data movement and offloads entirely to hardware Copy Engine DMA with zero SM occupancy. On 24 layers ($T=65{,}536$), bare zero-SM AllToAll runs slightly faster than AllGather plus ReduceScatter ($133.25$ vs.\ $138.65\text{ ms}$) while leaving $100\%$ of SMs free for overlapped compute and stream mixing.

3. \emph{Critical-path overlap and boundary behavior.} Under two-microbatch pipelining, each rank's $1/P$-scaled residual memory traffic is largely overlapped with the communication window, exposing only $+11.7\text{ ms}$ ($+3.0\%$) end-to-end. Removing FFN communication via replicated-weight pure SP confirms this boundary: without the FFN communication window, local stream traffic is directly exposed, raising \method{} overhead to $+6.8\%$ to $+7.6\%$ (with IHC at $+4.6\%$ to $+5.0\%$).

Overall, these results demonstrate that \method{}'s single-GPU staging overhead diminishes rapidly with model width and becomes minimal in realistic multi-GPU deployments where grouped branches map naturally onto tensor-parallel ranks, while further co-design between residual architectures and systems infrastructure can close the remaining single-device gap.